\documentclass[final]{siamltex}
\usepackage{algorithm,algpseudocode}
\usepackage{latexsym, graphicx, epsfig, amsmath, amsfonts,amssymb,bm}
\usepackage{caption, subcaption}
\usepackage{epstopdf}
\usepackage{booktabs}
\usepackage{array}
\usepackage{color}
\usepackage{url}
\usepackage{comment}

\newtheorem{remark}{Remark}[section]

\allowdisplaybreaks

\def\be{\begin{equation}}
\def\ee{\end{equation}}

\def\x{\mathbf{x}}
\def\y{\mathbf{y}}
\def\f{\mathbf{f}}

\def\N{\mathbf{N}}
\def\Nc{{{\mathcal N}}}

\def\PPh{\mathbf{\Phi}}
\def\PPs{\mathbf{\Psi}}

\def\R{{\mathbb R}}

\def\I{\mathbf I}
\def\M{\mathcal{M}}
\def\A{\mathbf{A}}
\def\D{\mathcal{D}}

\def\u{\mathbf{u}}
\def\v{\mathbf{v}}
\def\w{\mathbf{w}}
\def\U{\mathbf{U}}

\def\L{\mathcal{L}}

\title{Deep Neural Network Modeling of Unknown Partial Differential Equations in Nodal Space}

\author{Zhen Chen\footnotemark[1]\and Victor Churchill\thanks{Department of Mathematics,
		The Ohio State University, Columbus, OH 43210, USA. Emails:
		{\tt chen.7168@osu.edu, churchill.77@osu.edu, xiu.16@osu.edu.} Funding: This 
		work was partially supported by AFOSR FA9550-18-1-0102.}\and Kailiang 
		Wu\thanks{Department of Mathematics, Southern University of Science and Technology,
		Shenzhen 518055, P.R.China. Email: {\tt wukl@sustech.edu.cn.}}\and Dongbin Xiu\footnotemark[1]
				}

\begin{document}
\maketitle
\begin{abstract}
We present a numerical framework for deep neural network (DNN) modeling of unknown time-dependent partial differential equation (PDE) using their trajectory data. Unlike the recent work of [Wu and Xiu, J. Comput. Phys. 2020], where the learning takes place in
modal/Fourier space, the current method conducts the learning and modeling in physical space and uses measurement data as nodal values. 
We present a DNN structure that has a direct correspondence to the evolution operator of the underlying PDE, thus establishing the existence of
the DNN model.  The DNN model also does not require any geometric information of the data nodes. Consequently, a trained DNN defines a predictive
model for the underlying unknown PDE over structureless grids.
A set of examples, including linear and nonlinear scalar PDE, system of PDEs,  in both one dimension and two dimensions, over structured and unstructured grids, are presented to demonstrate the effectiveness of the proposed DNN modeling.
Extension to other equations such as differential-integral equation, is also discussed.
\end{abstract}

\section{Introduction} \label{sec:intro}

Data driven modeling and learning of unknown physical laws has
attracted much attention in recent years.
Most efforts were devoted to learning dynamical systems, i.e., systems
of ordinary differential equations (ODEs).
One popular approach is to construct a mapping from state variables
to their time derivatives. The mapping can be constructed as
a sparse approximation from a large dictionary set, such that
exact equation recovery can be achieved under certain circumstances. See, for example,
\cite{brunton2016discovering}, and its various
extensions
\cite{brunton2016discovering,kang2019ident, schaeffer2017sparse,schaeffer2017extracting,
  tran2017exact}. The approach has also been
applied to recovery of partial differential equations (PDEs) (\cite{rudy2017data,
  schaeffer2017learning}).
The mapping can also be constructed via the more
traditional approximation methods such as polynomial
approximation. This approach does not achieve exact equation
recovery but can ensure high accuracy in the more traditional sense.
See \cite{WuXiu_JCPEQ18}  and its extension to Hamiltonian systems
\cite{WuQinXiu2019}.
More recently, deep neural networks (DNNs) have been used to construct
the mapping. Studies of this direction include modeling of
ODEs
\cite{raissi2018multistep,qin2018data,rudy2018deep}, as well as
PDEs
\cite{long2017pde,raissi2017physics1,raissi2017physics2,raissi2018deep,long2018pde,sun2019neupde}.

A different approach for learning an unknown system is to construct a
mapping between two system states separated by a short time
(\cite{qin2018data}). For ODEs, the mapping is governed by the flow map of the
underlying system. Although not directly recovering the underlying equations,
the approach can define an accurate predictive model for the unknown
system, so long as an accurate approximation for the flow map is
constructed.
One practical advantage of the
flow map based approach is that it eliminates the need for temporal
derivative data, which can be difficult to acquire and may be
subject to larger errors.
The studies on this approach mostly rely on the use of DNNs.
In particular, residual network (ResNet),
developed in the image analysis community (\cite{he2016deep}), was found
to be highly suitable for recovering the flow map (\cite{qin2018data}). Since its introduction 
(\cite{qin2018data}), the flow map based DNN modeling approach has
been extended to modeling of non-autonomous 
dynamical systems (\cite{QinChenJakemanXiu_SISC}), parametric dynamical systems
(\cite{QinChenJakemanXiu_IJUQ}), partially observed dynamical systems
(\cite{FuChangXiu_JMLMC20}), as well as PDE
(\cite{WuXiu_modalPDE}).

The focus of this paper is on the development of a general numerical
framework for approximating/learning unknown time-dependent PDE. 
Even though the topic has been explored in several recent articles, cf.,
\cite{long2017pde,raissi2017physics1,raissi2017physics2,raissi2018deep,long2018pde,sun2019neupde},
the existing studies are relatively limited in scope, in the sense that they have mostly focused on 
 learning certain specific types of PDE or identifying some unknown terms in
 a given PDE.
 The contribution of this paper is in the development of a general DNN
 framework for modeling general types of unknown PDE. More
 specifically, instead of identifying the various terms in the unknown
 PDEs, as in most of the existing studies, our proposed method seeks to
 approximate the evolution operator of the underlying unknown
 PDE. This can be considered an extension of the flow map based
 learning for ODE
(\cite{qin2018data}) to PDE.
The evolution operator completely characterizes the time
 evolution of the PDE solution, and its accurate approximation allows
 one to construct an accurate predictive model for the underlying
 unknown PDE.
 The proposed method here is closely related to the
recent work of \cite{WuXiu_modalPDE}.
The main difference is that the method of \cite{WuXiu_modalPDE} conducts
the learning in modal space, i.e., generalized Fourier space, and
requires a set of pre-defined basis functions and transformation of
measurement data into the modal space. Our current work
conducts the learning directly in physical space by using
measurement of the solution states on a set of grids/nodes. To achieve
this, we propose
a new DNN structure, consisting of a disassembly block and an assembly
layer, that has a direct correspondence to a general time-stepping
evolution of the unknown PDE. Subsequently, the mathematical existence of the DNN
model for the unknown PDE can be established. On the practical side, the proposed DNN
structure allows one to use structure-free grids/nodes without any
geometric information. This can be advantageous for many practical
problems, where measurement data are collected at rather arbitrary
physical locations.
A comprehensive set of
examples are presented to demonstrate the effectiveness of the proposed
DNN learning. The examples include learning of linear
and nonlinear scalar PDE, systems of PDEs, in both one spatial
dimension and two spatial dimensions, over structured and unstructured
grids. The proposed DNN framework can also be used to learn other
types of equations, e.g., differential-integral equation.

This paper is organized as follows. After the problem setup and
preliminary materials in
Section \ref{sec:setup}, we present the main approach 
in Section
\ref{sec:method}, which includes the mathematical justification in
Section \ref{sec:math} and our DNN structure in Section \ref{sec:DNN}.
Upon discussing the model training in Section \ref{sec:training}, we
discuss the
extensions to system of PDEs in Section \ref{sec:system} and other
differential equations in Section \ref{sec:others}. Finally, we present
numerical examples in
Section \ref{sec:examples} to
demonstrate the properties of the proposed approach.

\section{Preliminaries} \label{sec:setup}

We first introduce the problem setup and notation for our
discussion. For notational clarity, we start the discussion on scalar
PDE. Extension to systems of PDEs is straightforward and will be left to
the latter part of this paper.

\subsection{Problem Setup}

Let us consider 
an autonomous time-dependent PDE,
\be \label{govern}
\begin{cases}
u_t = \L(u), \quad &(x,t) \in \Omega \times \mathbb R^+,
\\
{\mathcal B} (u) = 0, \quad &(x,t) \in \partial \Omega \times \mathbb
R^+,\\
u(x,0) = u_0(x), \quad & x \in \bar{\Omega}, 
\end{cases}
\ee
where
 $\Omega\subset \R^d$, $d=1,2,3$, is the physical domain under consideration, and ${\mathcal L}$ and ${\mathcal B}$
stand for the PDE operator and boundary
condition operator, respectively. Our basic assumption is that the PDE is unknown.

We assume data of the state variable $u$ are available over a set of nodal points, or
grids,
$$
X_N = \{x_1,\dots, x_N\} \subset \Omega,
$$
and by using vector notiation, we write
$$
\u(t) = (u(x_1,t),\dots, u(x_N,t))^T.
$$
Also, the data are only available at certain discrete time instances,
resulting in so-called snapshots of the solution
$$
\u(t_j^{(k)}), \qquad j=1,\dots, \ell^{(k)},
\quad k=1,\dots, N_{traj}.
$$
Here the superscript $k$ denotes the $k$-th ``trajectory'', which
implies all $\ell^{(k)}$ snapshots are evolved from
the same (unknown) initial state, and $N_{traj}$ denotes the total
number of such ``trajectories''.
We then
group the solution snapshots into pairs at two consecutive time
instances,
$$
( \u(t_j^{(k)}), ~~\u(t_{j+1}^{(k)})), \qquad j=1,\dots, \ell^{(k)}-1,
\quad k=1,\dots, N_{traj}.
$$
For autonomous PDE, the time variable can be
arbitrarily shifted. After collecting all the data pairs over all the
trajectories and re-ordering them using a single index $m$, we obtain
\be \label{data-pair}
( {\u}_m(0), {\u}_m(\Delta t)), \qquad m=1,\dots, M,
\ee
where $\Delta t>0$ is the time step between each pair, and $M=
\ell^{(1)}+\cdots+\ell^{(N_{traj})}-N_{traj}$ the total number of
data pairs. This data set \eqref{data-pair} shall be our training data
set, where we assumed $\Delta t$ as a constant time step only for
notational convenience. (Variable time steps can be readily modeled as
an additional input parameter, as discussed in \cite{QinChenJakemanXiu_IJUQ}.)
Each pair in the data set can be considered as a solution trajectory of
two entries, starting with the ``initial condition'' $\u_m(0)$ and
ending with the state $\u_m(\Delta t)$ a short time $\Delta t>0$ later.

Our goal is to construct an accurate approximation of the evolution/dynamics of the unknown
governing equation \eqref{govern} via the snapshot data
\eqref{data-pair}. Once the approximation is constructed, it can serve
as a predictive model to provide prediction and analysis of the
unknown PDE system.

\subsection{Deep Learning of ODE Systems}

We now briefly review flow-map based deep learning of ordinary
differential equations (ODEs) \cite{qin2018data}. The idea behind the flow-map
based deep method is used as a starting point of our method for PDE
learning.

Consider an unknown dynamical system
$d\x/dt = \f(\x)$, where $\x\in\R^n$ and $\f$ is unknown.
Let $\PPh:\R^{n}\times\R\to\R^{n}$ be the flow map of the system.
The state variables satisfy
$
\x(s_1) = \PPh(\x(s_0),s_1-s_0),
$
for any $s_1\geq s_0$.
Suppose data of the state
variables $\x$ are available as
$$
 \x\left(t^{(k)}_j\right), \qquad j=1,\dots, \ell^{(k)},
 \quad k=1,\dots, N_{traj},
$$
for a total number of $N_{traj}$ trajectories over time instances
$\{t_j^{(k)}\}$ with step size
$\Delta_j^{(k)}=t_{j+1}^{(k)}-t_{j}^{(k)}$.

For the flow-map based learning method developed in \cite{qin2018data},
we first re-group the trajectory data 
as pairs of two
adjacent time instances,
 \be \label{dataset0}
 \left\{\x_m(0), \x_m(\Delta)\right\}, \qquad m=1,\dots, M,
\ee
where $M$ is the total number of such data pairs, and for notational
convenience the time step $\Delta_j^{(k)}$ is
assumed to be a constant $\Delta$ for all $j$ and $k$.
Note that for autonomous system, time $t$ can be
 arbitrarily shifted and only the relative time difference is
 relevant. 

 We then construct a standard fully connected feedforward
 deep neural network, whose input and output layers both have $n$
 neurons. Let $\N:\R^n\to\R^n$ be its associated
 mapping operator and define residual network (ResNet) mapping
 (\cite{he2016deep})
     \begin{equation}\label{resnet}
      \y^{out} =  \left[\mathbf{I}_n +
        \N\right]\left(\y^{in}\right),
    \end{equation}
    where $\mathbf{I}_n$ is the identity matrix of size $n\times n$.
    Upon using the data set
    \eqref{dataset0} and letting $\y^{in}\leftarrow \x(0)$ and
    $\y^{out}\leftarrow \x(\Delta)$, the network operator $\N$ can be
    trained using mean square loss, such that
    $$
    \x_m(\Delta) \approx \x_m(0) + \N(\x_m(0)),
    $$
    where the approximation error is controlled by the training
    algorithm.
%
    Upon satisfactory training
of the network, we then obtain a predictive model
$$
\x(t_{j+1}) = \x(t_j) + \N(\x(t_j)), \qquad j=0,\dots,
$$
for any arbitrarily given initial condition $\x(t_0)$. Properties of
the flow-map based learning were discussed in \cite{qin2018data}, with
its extension for variable
time step $\Delta_j^{(k)}$ and other system parameters in
\cite{QinChenJakemanXiu_IJUQ} and systems with missing
variables in \cite{FuChangXiu_JMLMC20}.

\subsection{Learning of PDE in Modal Space}

In \cite{WuXiu_modalPDE},  the flow-map based deep learning method was extended to
PDE learning. The key is to conduct the learning in modal space, i.e.,
generalized Fourier space. Consider $u:\Omega\times [0,T]\to \R$, $T>0$,
as the solution field of an unknown PDE
$$
u_t = \L(u).
$$
One first chooses a set of basis functions in the physical domain $\Omega$,
$\{b_j(x)$, $j=1,\dots, N_b\}$ and expresses the solution as a finite-term
series
$$
u(x,t) = \sum_{j=1}^{N_b} \hat{u}_j(t) b_j(x),
$$
where $\hat{\u}(t) = [\hat{u}_1(t), \dots, \hat{u}_{N_b}(t)]^T$ are the
expansion coefficients. Thus there exists a system of ODEs for the
expansion coefficients, in the form of
$$
\frac{d\hat{\u}}{dt} = \f(\hat{\u}).
$$
If the governing PDE equation is known, such a system for $\hat{\u}$ can be
derived via a numerical approximation technique, e.g., Galerkin
method. When the governing PDE is unknown, the system for $\hat{\u}$ is
unknown as well. When solution data are available, this unknown ODE
system can be learned in the modal space by extending
flow-map based method from \cite{qin2018data}.
Details of the modal space PDE learning, including its proper
mathematical formulation and data
processing procedure, can be found in \cite{WuXiu_modalPDE}.

\section{DNN Modeling Approach} \label{sec:method}

In this section, we present our DNN modeling of unknown PDE in nodal
space. The method allows direct use of measurement data on grid points
and does not require structural information of the grids.
We first
present the basic mathematical formulation, which motivates
the base design of our DNN structure.
We then discuss the important properties of the proposed DNN network, as well
as its extension to learning of system of PDEs and other differential equations.

\subsection{Mathematical Motivation} \label{sec:math}

Let $\alpha\in \mathbb{N}_0^d$ be $d$-variate multi-index with
$$
\alpha = (\alpha_1, \dots, \alpha_d), \qquad
|\alpha| = \alpha_1+\cdots+\alpha_d.
$$
Without loss of generality, we  consider a $p$-th order autonomous PDE in the
following general form,
\be \label{PDE}
u_t = \L(u, \partial^{(1)} u, \dots \partial^{(p)} u), \qquad
p\geq 1,
\ee
where, for any $1\leq m\leq p$,
$$
\partial^{(m)} = \{\partial_{x_1}^{\alpha_1}\cdots
\partial_{x_d}^{\alpha_d}: |\alpha|=m\}.
$$
Note that for multi-dimensional PDE with $d>1$, each $\partial^{(|\alpha|)}$
represents multiple partial derivative operators. Also,
$u=\partial^{(0)} u$. Without considering boundary conditions, we
assume the PDE \eqref{PDE} (1) is well-posed; and (2) has finite
number $N_D\geq p$ of partial derivative terms.

Let $X_N = \{x_1,\dots, x_N\}\subset \Omega$ be the grid set, and
$\mathcal{T} = \{t^k
= k \Delta t$, $k=0,\dots, K\}$ be uniformly distributed (for notational
convenience) time
instances up to $T=K\Delta t$. We assume that when the PDE \eqref{PDE} is known, it can be
approximated by a one-step Euler forward type 
explicit numerical scheme with sufficiently small local truncation
error. More specifically, let
\be \label{scheme}
\v_N^{k+1} = \v_N^k + \Delta t \cdot \PPs(\v_N^k), \qquad k=0,\dots, K-1,
\ee
be the numerical scheme,
where $\v_N^k = (v(x_1,t^k),
\dots, v(x_N, t^k))^T$ is the numerical solution over $X_N$ at time
$t^k$, and $\PPs$ is the incremental function.
We assume that
the grid set $X_N$ admits a proper spatial
discretization of the PDE \eqref{PDE} such that
\be \label{consistency}
\u_N^{k+1} = \u_N^k + \Delta t \cdot (\PPs(\u_N^k)
+ {\bm \tau}_N^{k+1}), \qquad k=0,\dots, K-1,
\ee
where $\u_N^k = (u(x_1,t^k),\dots, u(x_N,t^k))^T$ is the exact solution
on $X_N$ at time $t^k$, and
${\bm \tau_N^{k+1}}$ is the local truncation error satisfying
\be \label{LTE}
\epsilon(X_N, \Delta t) = \max_k \|{\bm \tau}_N^{k+1}(X_N,\Delta t)\|  \ll 1.
\ee
Note that this is a very mild assumption. We only assume that if the
PDE \eqref{PDE} is known, it can be discretized over the grid set
$X_N$ and time domain $\mathcal{T}$ with Euler-forward, such that the
global truncation error is sufficiently small. The error obviously
depends on the choice of spatial discretization over $X_N$
and scales as $O(\Delta t)$ because of Euler-forward temporal
discretization.
We do not assume
the stability of the scheme.

\begin{proposition} \label{prop1}
Under the assumption that the PDE \eqref{PDE} admits the Euler forward
explicit numerical method \eqref{scheme} satisfying \eqref{LTE}, there exists a
set of functions $\{F_i:\R^N\to\R^N, i=1, \dots, J\}$, $J\geq 1$, and
an iterative scheme 
 \be \label{basic}
\v_N^{k+1} = \v_N^k + \Delta t \cdot \M\left[F_1(\v_N^k), \cdots,
F_J(\v_N^k)\right],  \qquad k=0,\dots, K-1,
\ee
where $\M$ is a (nonlinear) function operated component-by-component,
such that for sufficiently large $J$, the exact solution of \eqref{PDE} satisfies
\be \label{basic1}
\u_N^{k+1} = \u_N^k + \Delta t \cdot \left(\M\left[F_1(\u_N^k), \cdots, F_J(\u_N^k)\right]
+ {\bm \eta}_N^{k+1}\right), \qquad k=0,\dots, K-1,
\ee
where $\|{\bm \eta}_N^{k+1}\| \leq \epsilon (X_N, \Delta t) $,
the global truncation error defined in \eqref{LTE}, $\forall k$.
\end{proposition}
\begin{proof}
  The existence of \eqref{basic} can be proved by
  construction, based on
  the explicit Euler-forward scheme \eqref{scheme}.
  Given the grid set $X_N$, there exist linear operators to
  approximate partial derivatives. More specifically, consider a sufficiently
  smooth function $w(x)$ and its grid function $w(x)|_{X_N}=
  (w(x_1),\dots, w(x_N))^T:=\w$.
  Then, its partial derivatives on $X_N$ can be approximated via
  differentiation matrices.  That is,
for any $\alpha$ with $1\leq |\alpha|\leq p$, there exists a
differentiation matrix $\mathbf{D}^\alpha:\R^N\to \R^N$ such that,
\be\label{Diff}
 \partial_{x_1}^{\alpha_1}\cdots
 \partial_{x_d}^{\alpha_d} w(x) |_{X_N} \approx \mathbf{D}^{\alpha} \w.
 \ee

Upon using Euler forward time stepping over $\mathcal{T}$, one can
construct the following explicit numerical approximation to the PDE 
\eqref{PDE}.
  \be \label{Euler}
  \v_N^{k+1} = \v_N^k + \Delta t \cdot \L\left(\v_N^k,
\mathbf{D}^{(1)}\v_N^k, \cdots, \mathbf{D}^{(p)}\v_N^k\right), \qquad
k=0,\dots, K-1,
\ee
where
$$
\mathbf{D}^{(m)} = \{ \mathbf{D}^\alpha: |\alpha|=m\}, \qquad 1\leq m
\leq p.
$$
%
%
This is a special case of \eqref{basic}, where $F_1$ is the identity
operator, $F_2,\dots, F_J$ correspond to the differentiation
matrices in \eqref{Euler}, and $J=N_D+1$. Note $N_D$ is the number of
partial derivatives in \eqref{PDE}.  The existence of \eqref{basic} is thus established.
  \end{proof}

  \begin{remark}
    Constructing differentiation matrices over a grid set $X_N$, in
    the form of \eqref{Diff}, is a standard procedure in 
    collocation type numerical PDE methods, e.g., finite difference method,
    spectral collocation method, radial basis method, etc.
    The accuracy of the differentiation matrix approach depends
    primarily on the property of the grid $X_N$, the chosen
    approximation approach, and the regularity of the function under
    consideration.
\end{remark}

\begin{remark}
The accuracy of the Euler forward linear approximation scheme
\eqref{Euler}, manifested by its truncation error \eqref{LTE},
is obviously limited. In particular, its temporal accuracy is limited
to first order $O(\Delta t)$. However, the scheme \eqref{Euler} is only needed
to establish the existence of \eqref{basic}, which is of more general
nonlinear form. The nonlinear scheme \eqref{basic} serves as the
foundation of our deep learning algorithm, which will not be restricted to
by the accuracy of the Euler forward method \eqref{Euler}. 
\end{remark}

\subsection{Basic DNN Structure} \label{sec:DNN}

Deep neural networks (DNNs) are naturally nonlinear approximators.
In
this section, we discuss our DNN structure design to achieve the nonlinear
approximation, in
the form of \eqref{basic}, of an unknown PDE \eqref{PDE}.
Once successfully trained via the available data set
\eqref{data-pair}, the DNN model provides an iterative scheme, as in
\eqref{basic}, that can be used as a predictive model of the
underlying unknown PDE.

\begin{figure}[htbp]
	\begin{center}
		\includegraphics[width=15cm]{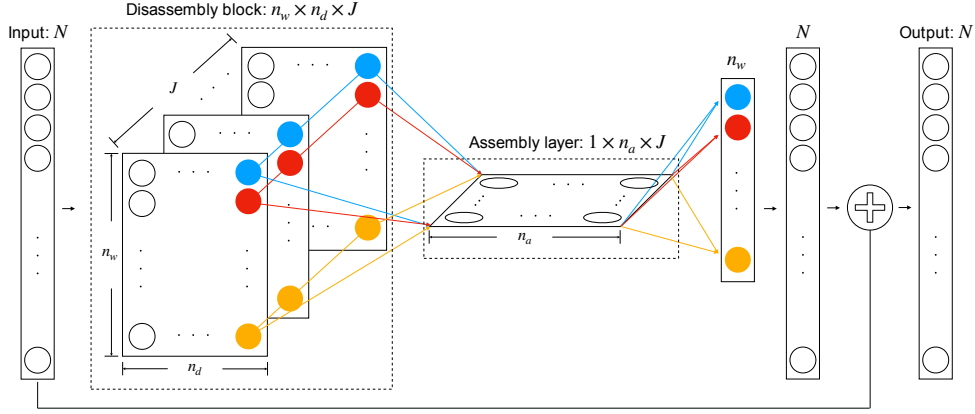}
		\caption{The basic DNN structure.}
		\label{fig:basic_NN}
	\end{center}
\end{figure}

The basic structure of our DNN model is illustrated in Figure \ref{fig:basic_NN}.
It consists of the following components:
\begin{itemize}
\item Input layer, where the number of neurons is $N$, the dimensionality
  of $\u$.
\item Disassembly block. It has $J\geq 1$ fully connected FNNs
  ``in parallel'', where each FNN has width $n_w$ and depth $n_d$ and
  receives inputs from the input layer. As
  illustrated in Figure \ref{fig:basic_NN}, the disassembly
  block thus creates a 3-dimensional tensor structure with dimension
  $n_w\times n_d\times J$, where $J$ shall be referred to as the
  ``thickness'' of the disassembly block hereafter. The output
  layers of the disassembly block create a matrix structure of $n_w\times
  J$, spanning the width and thickness directions of the block.
\item Assembly layer. It is a standard fully connected FNN, with width
  $J$ and depth $n_a$. It
  operates along the thickness direction of the disassembly
  block. It receives input from
  one of the ``rows'' of  
  the disassembly block output matrix, whose size is $n_w\times J$, and produces a scalar output. This is
  repeated for each of the $n_w$ rows of the disassembly output matrix
  and produces an output vector of $n_w$. (In other words, one can
  also view the assembly layer as a block of $n_w$ identical FNNs, i.e.,
  with shared parameters,
  stacked vertically.)
\item Output layer,  where the number of neurons is $N$, the dimensionality
  of $\u$. The output layer is mapped from the output of the assembly
  layer, whose dimension is $n_w$. The input layer is re-introduced
  before the final output, in the same manner of residual network (ResNet).
\end{itemize}

Let $\Nc_i$, $i=1,\dots, J$, be the operators defined by the
parallel FNNs inside the disassembly block, i.e.,
\be
\Nc_i:\R^N\to \R^{n_w}, \quad i=1,\dots, J.
\ee
Let $\D$ denote the operator defined by the disassembly
block. We then have,
with a slight abuse of notation,
\be
\D = \bigotimes_{1\leq i\leq J} \Nc_i: \R^N\to \R^{n_w\times J}.
\ee
Let $\A$ be the operator defined by the assembly layer. We have
\be
\A: \R^J\to \R,
\ee
which is applied to each ``row'' of the output of the disassembly
block. The ``component-wise'' application of $\A$ results in the
following operator composition:
\be \label{AD}
\A\circ\D: \R^N\to \R^{n_w}.
\ee
Let
\be \label{F}
\mathcal{F}:\R^{n_w}\to\R^N
\ee
be an operator mapping between $n_w$-vector and $N$-vector. We can then write the operator for the
entire network structure, shown in Figure \ref{fig:basic_NN}, as
\be
{\bm \Nc} = \I_N + \mathcal{F}\circ\A\circ\D :\R^N\to \R^N,
\ee
where $\I_N$ is the identity matrix of $N\times N$.
The proposed neural network thus defines the following operation:
for $\w^{in}\in\R^N$ and $\w^{out}\in\R^N$,
\be \label{NN_map}
\w^{out} = {\bm \Nc}(\w^{in}; \Theta) = \w^{in}+ \mathcal{F}\left[\A(\Nc_1(\w^{in}), \dots, \Nc_J(\w^{in}))\right],
\ee
where $\Theta$ is the set of network parameters and $\A$ is operated
component-wise.

\subsection{Model Properties} \label{sec:model}

Let us first consider a special case of the proposed DNN model, where the
width of each of the FNN in the disassembly block $n_w = N$, the
dimensionality of the input. In this case, the output of the assembly
layer also has dimension $N$, i.e., from \eqref{AD} we have
\be
\A\circ\D: \R^N\to \R^{N}.
\ee
Subsequently, the $\mathcal{F}$ operator \eqref{F} can take the form
of the 
identity operator $\I_N$. The corresponding DNN model \eqref{NN_map}
becomes
\be \label{NN_simple}
\w^{out} = \w^{in}+ \A(\Nc_1(\w^{in}), \dots, \Nc_J(\w^{in})).
\ee
This is a direct correspondence to the mathematical model
\eqref{basic} in Proposition \ref{prop1}.  Each FNN in the disassembly block approximates
the $F$'s function in \eqref{basic}, 
\be
\Nc_i \approx F_i, \qquad i=1,\dots, J,
\ee
and the assembly layer approximates the $\M$ operator in \eqref{basic},
\be
\A\approx \M.
\ee

Courtesy of the universal approximation property of neural networks (cf. \cite{scarselli1998universal}),
the DNN model \eqref{NN_simple} thus can approximate the general
mathematical model \eqref{basic} arbitrarily well, when all the FNNs
involved in the network have sufficient complexity.

The general case of $n_w\neq N$ is worth discussing, especially the
case of $n_w < N$. In many problems, the dimensionality of the grid
set $X_N$ can be exceedingly large, i.e., $N\gg 1$. 
Adopting $n_w<N$, and in some cases, $n_w \ll N$, can be
computationally advantageous. In this case, the core of the proposed DNN model,
which consists of the disassembly block and assembly layer, performs
\eqref{AD} and achieves effectively a model reduction
operation $\R^N\to \R^{n_w}$. Subsequently, the $\mathcal{F}$ operator
\eqref{F} achieves 
a ``lifting'' operation back to the higher dimension spacer
$\R^{n_w}\to\R^N$. This inherent model reduction capability of our DNN
model is worth noting. This property and its analysis are beyond the scope
of this paper and will be pursued in a separate study. In this paper, we
will focus on the special case of $n_w = N$. 

Another notable property of our DNN model is that both the input
vector $\w^{in}$ and output vector $\w^{out}$ can be arbitrarily
permuted, so long as they are subjected to the same permutation. Since the
indexing of these vectors follow the grid set $X_N$, an immediate
consequence is that the DNN model requires no geometric information of
the underlying grid set $X_N$. Consequently, our DNN model can work
with arbitrarily distributed grid $X_N$ and results in a mesh-free
predictive model.

\subsection{Model Training and Prediction} \label{sec:training}

Our nodal PDE learning model is constructed by training the DNN
model  \eqref{NN_map} with the data set
\eqref{data-pair}. Upon setting
\be
\w^{in} \leftarrow \u_j(0), \qquad j=1,\dots, M,
\ee
we minimize the mean squared loss
\be \label{loss1}
L(\Theta; \Delta t) = \frac{1}{M}\sum_{j=1}^M \left\|  {\bm \Nc} (
  {\u}_j(0); \Theta)   - \u_j(\Delta t) \right\|_2^2
\ee
to obtain the network hyperparameters $\Theta$.

Once trained, we obtain a predictive model over the grid set $X_N$ for the underlying unknown
PDE. For an arbitrarily given initial condition $\u_N(0)$ over the grid $X_N$, we
have
\be \label{NN_model}
\left\{
  \begin{split}
 &   \v_N^0 = \u_N(0), \\
&\v_N^{k+1} = {\bm \Nc}(\v^{k}) = \v_N^{k}+ \mathcal{F}\left[ \A(\Nc_1(\v_N^{k}), \dots,
\Nc_J(\v_N^{k}))\right], \qquad k=0,\dots.
\end{split}
\right.
\ee

Although this DNN model resembles the Euler forward type of numerical
methods, it is fundamentally different. The traditional numerical
methods, including Euler forward, are defined through predetermined
coefficients and parameters. They are usually linear methods and applicable to general 
PDEs. The DNN model \eqref{NN_model} is fully nonlinear, in the sense
that it is constructed specifically for the PDE system behind the
training data. As a result, the DNN model does not have
temporal error associated with $\Delta t$, as in the traditional
numerical methods (cf. \cite{qin2018data}). On the other hand, the
trained DNN model can only
be applied to prediction of the PDE associated with the training data and
not to another PDE system.

Numerical stability of the predictive model \eqref{NN_model} can be
not fully analyzed at the moment, due to the lack of mathematical
tools for DNNs. However, we have discovered that, upon extensive numerical
experimentation,
numerical stability of the DNN model can be drastically
improved by using recurrent loss, i.e., loss computed over multiple
time steps. Let $n_L\geq 1$ be the number of time steps for loss
computation, our loss function is defined as
\be \label{loss_n}
L(\Theta) = \sum_{n=1}^{n_L} L(\Theta; n\Delta t),
\ee
where, for $n=1,\dots, n_L$,
$$
L(\Theta; n\Delta t) = \frac{1}{M}\sum_{j=1}^M \left\|  {\bm \Nc}^n (
  {\u}_j(0); \Theta)   - \u_j(n\Delta t) \right\|_2^2,
$$
with ${\bm \Nc}^n={\bm \Nc}\circ\cdots\circ {\bm \Nc}$ composed $n$
times. Note that the mean squared loss defined in \eqref{loss1} is
merely a specical case of $n_L=1$. The use of the recurrent loss with
$n_L>1$ does require us to re-orgainize the training data set
\eqref{data-pair}. In this case, the training data set needs to
contain data sequences of length $n_L+1$.
\be \label{data-set}
\left( {\u}_m(0), {\u}_m(\Delta t), \dots, \u_m(n_L\Delta t)\right), \qquad m=1,\dots, M.
\ee
Again, the original data set
\eqref{data-pair} is a special case of $n_L=1$.

\subsection{Extensions}  \label{sec:extension}

\subsubsection{PDE Systems} \label{sec:system}

The proposed DNN model can be extended to system of PDEs in a
straightforward manner. For a system of PDEs with state variables
$U=(u_1,\dots, u_L)$ and governed by unknown PDE system
\be
\partial_t u_i = \mathcal{L}_i(U), \qquad i=1,\dots, L,
\ee
where the PDE operator $\mathcal{L}_i$ includes partial derivatives of
$U$. Our DNN model \eqref{NN_map} for scalar PDE \eqref{PDE}, as illustrated in Figure
\ref{fig:basic_NN}, can be extended to the system by merely
concatenating all the state variables into a single larger
vector. That is,
let $\u_i\in \R^N$ be the data snapshot of each component $u_i$, $i=1,\dots,
L$. We then define
\be
\U = [\u_1^T, \cdots, \u_L^T]^T\in \R^{N L}
\ee
as the concatenated state vector. The DNN structure and network model
in the previous sections can then be applied directly to $\U$.

Note that it is possible, and maybe even preferable mathematically, to
construct a more ``refined'' DNN structure to apply to each of the
PDEs in the system. However, our early numerical experimentation
indicated that the straightforward concatenation approach is a
flexible numerical approach. Further refinement in DNN structure for
PDE systems will be investigated in a future study.

\subsubsection{Other Differential-type Equations} \label{sec:others}

The proposed DNN structure in Section \ref{sec:DNN} has its foundation
in Section \eqref{sec:math}, particularly in Proposition
\ref{prop1}. The key ingredient is that partial derivatives can be
approximated by the solution values over the grid $X_N$. This is
explicitly achieved in the standard numerical methods via techniques
such as
differentiation matrices; and implicitly achieved in our DNN model
via the disassembly block structure. A natural extension is that operators other
than partial differentiation can also be incorporated and modeled by
the same DNN structure, as long as the operator can be approximated
using solution values over the grid $X_N$. One obvious example is
integral operator over the domain $\Omega$. More specifically, one can
always use data over the grid $X_N$ to approximate an integral, i.e.,
\be \label{integral}
\int_\Omega u(x) dx \approx G(u(x_1),\dots u(x_N)).
\ee
The existence of $G$ is obvious, as there are a variety of integration
rules (e.g, quadrature, sampling methods) to accomplish the
approximation. Therefore, if such an integral exists in the PDE, the same DNN structure in Section
\ref{sec:DNN} can be applied. The applicability of the proposed DNN
model to differential-integral equations will be demonstrated in our
numerical example section.
\section{Computational Studies} \label{sec:examples}

In this section, we present numerical examples to demonstrate the properties
of the proposed DNN modeling method for PDEs. In all of the examples,
the true governing PDEs are known. These true PDEs serve only two
purposes: (1) to generate synthetic training data. This is achieved by
solving the true PDEs either analytically (if possible) or by high
resolution numerical solvers (most of the cases). Once the training
data sets are generated, they are used to train DNN models for
the PDEs behind the data; and (2) to generate reference solutions for
validating the DNN model predictions. 
Therefore, the knowledge of the true PDEs does
not in any way facilitate the DNN model construction. Also, the
training data sets \eqref{data-set} consist of very short trajectories of length
$n_L\times \Delta t$, where $n_L$ is the number of time steps in
computing the loss function. In all of the examples, we have $n_L\leq 10$. For predictions, we typically march the DNN models
under arbitrarily given new initial conditions
that are not in the training data set for a much longer time
horizon. Therefore, all our DNN model predictions are
``out-of-sample''.

The one-dimensional examples have periodic boundary conditions, and the
two-dimensional example has Dirichlet boundary conditions. Note that
the prescribed boundary conditions are used only to
generate the training data. The boundary conditions are not enforced
in the DNN modeling structure. Instead, the DNN models ``learn'' the
correct boundary conditions embedded in the training data and proceed
to produce predictions over longer time.

For the one-dimensional examples with periodic boundary conditions, we
generate training data by solving the true PDEs with the following
``initial conditions'', in the form of finite Fourier series,
\begin{equation} \label{Fourier}
	u(x,0) = a_0+ \sum_{n=1}^{N_c} \left ( a_n\cos(nx) +
          b_n\sin(nx) \right ),
\end{equation}
where the Fourier coefficients $\{a_n, b_n\}$ are randomized to generate
different initial conditions. The training data sets are obtained by
solving the true PDEs for $n_L$ steps, as shown in \eqref{loss_n}, from
these initial conditions. In most of our examples, we set $n_L=10$.

Our numerical experimentations, along with the earlier work on flow
map based learning (\cite{qin2018data, QinChenJakemanXiu_IJUQ,
  WuXiu_modalPDE, FuChangXiu_JMLMC20}, indicate that one does not need
exceedingly complex DNNs for accurate modeling.
Therefore, in the following examples, the 
disassembly block and assembly layer have depths $n_d = 1$ and $n_a = 1$, 
while the width and thickness of the disassembly block are set as $n_w=N$ and $J=5$. Hence, 
unless otherwise noted, our DNNs have disassembly 
block of dimension (width $\times$ depth $\times$ thickness) 
$N \times 1 \times 5$, and assembly layer of dimension $1\times 1
\times 5$, as illustrated in Fig. \ref{fig:basic_NN}.

\subsection{Advection-Diffusion Problem}

We first consider an advection-diffusion equation with $2\pi$-periodic boundary condition,
\begin{equation} \label{adv_diff}
 \frac{\partial u(x,t)}{\partial x} + \frac{\partial}{\partial x}(\alpha(x)u(x,t)) = \frac{\partial}{\partial x}\left(\kappa(x)\frac{\partial u(x,t)}{\partial x}\right), 
\end{equation}
where $\alpha$ is transport velocity and $\kappa$ is diffusivity. Note that both $\alpha$ and $\kappa$ are functions. Without
loss of generality, we set the functions as deterministic functions, in the Fourier series form \eqref{Fourier}, with
(arbitrarily chosen) fixed coefficients listed in Table \ref{table:coef}. Upon setting up the true governing equation, we use it
to generate training data. The data are generated by solving the equation with a high-resolution numerical solver --
Fourier collocation in space and Crank-Nicolson in time. We employ $10,000$ training data trajectories, each of
which are generated by solving the true equation with a randomized initial condition \eqref{Fourier} with $a_0=0$,  $a_n$ and $b_n$ drawn from a continuous uniform distribution over $[-1,1]$, i.e. $a_n,b_n\in U[-1,1]$, and $N_c$ chosen from a discrete uniform distribution over $\{1,\dots, 10\}$, i.e. $N_c\in U\{1,\dots,10\}$. The training trajectory data have time step $\Delta t=0.02$ and length $n_L=10$ in the loss
function \eqref{loss_n}. This implies that our training data are trajectories of length only $t=0.2$.
\begin{table} [htbp]
\begin{center}
\begin{tabular}{| c | c | c |}
  \hline
  $\alpha(x)$ & $a_0$ &1   \\
            & $a_n$ & $(0.5426, 0.2673, -0.0030, -0.6039, 0.6618)\times 0.05$  \\
              & $b_n$ & $(-0.9585, 0.4976, -0.5504, 0.5211, -0.8233)\times 0.05$ \\
  \hline
  $\kappa(x)$ & $a_0$  & $1\times 10^{-3}$ \\
            & $a_n$ & $(0.3707, -0.9921, 0.62524, 0.4435, 0.8355)\times 5\times 10^{-5}$  \\
              & $b_n$ & $(0.9068, 0.0243, 0.2251, -0.4162, 0.4291)\times 5\times 10^{-5}$ \\
\hline
\end{tabular}
\caption{The fixed parameters for $\alpha(x)$ and $\kappa(x)$ in \eqref{adv_diff}, in the form of \eqref{Fourier}.}
\label{table:coef}
\end{center}
\end{table}

Hyperbolic tangent function ``$\tanh$'' is used as the activation function throughout.
A cyclic learning rate is used during training, with maximum learning
rate of $1\times10^{-3}$, minimum learning rate of $1\times10^{-4}$,
and exponential decay rate of $0.99994$.
We trained the DNN for $5,000$ epochs with batch size 50.

\begin{figure}[htbp]
	\begin{center}
		\includegraphics[width=0.49\textwidth]{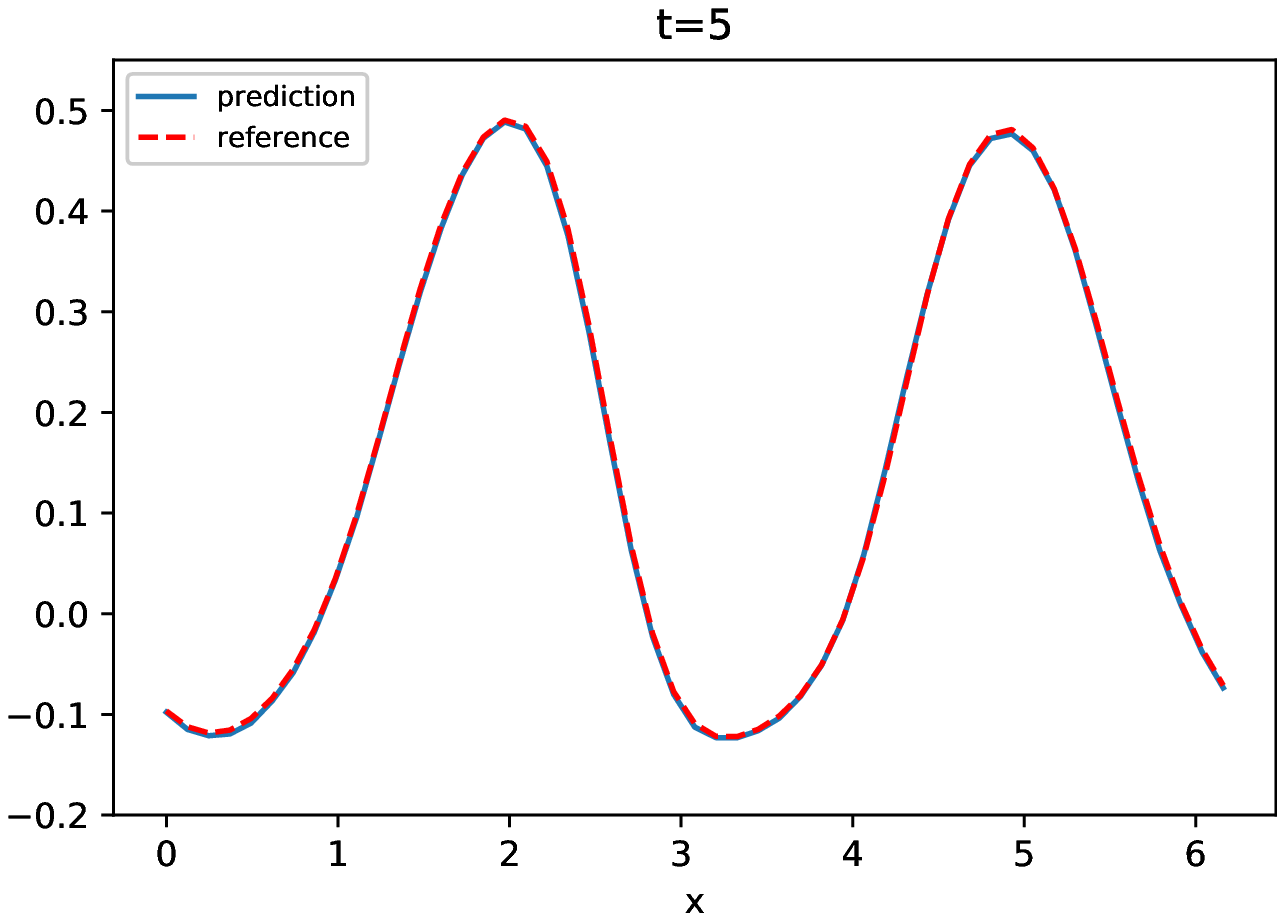}
		\includegraphics[width=0.49\textwidth]{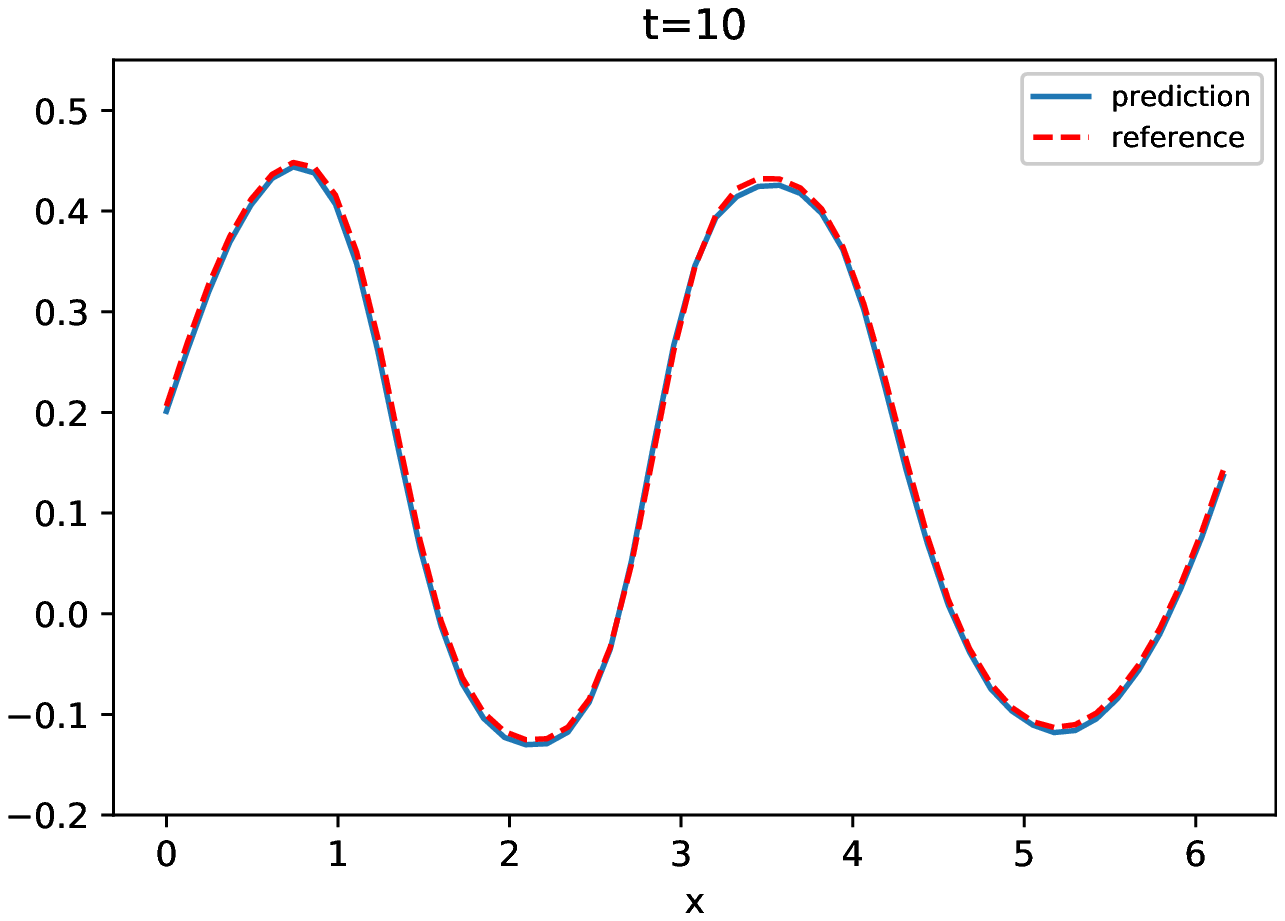}
		\includegraphics[width=0.49\textwidth]{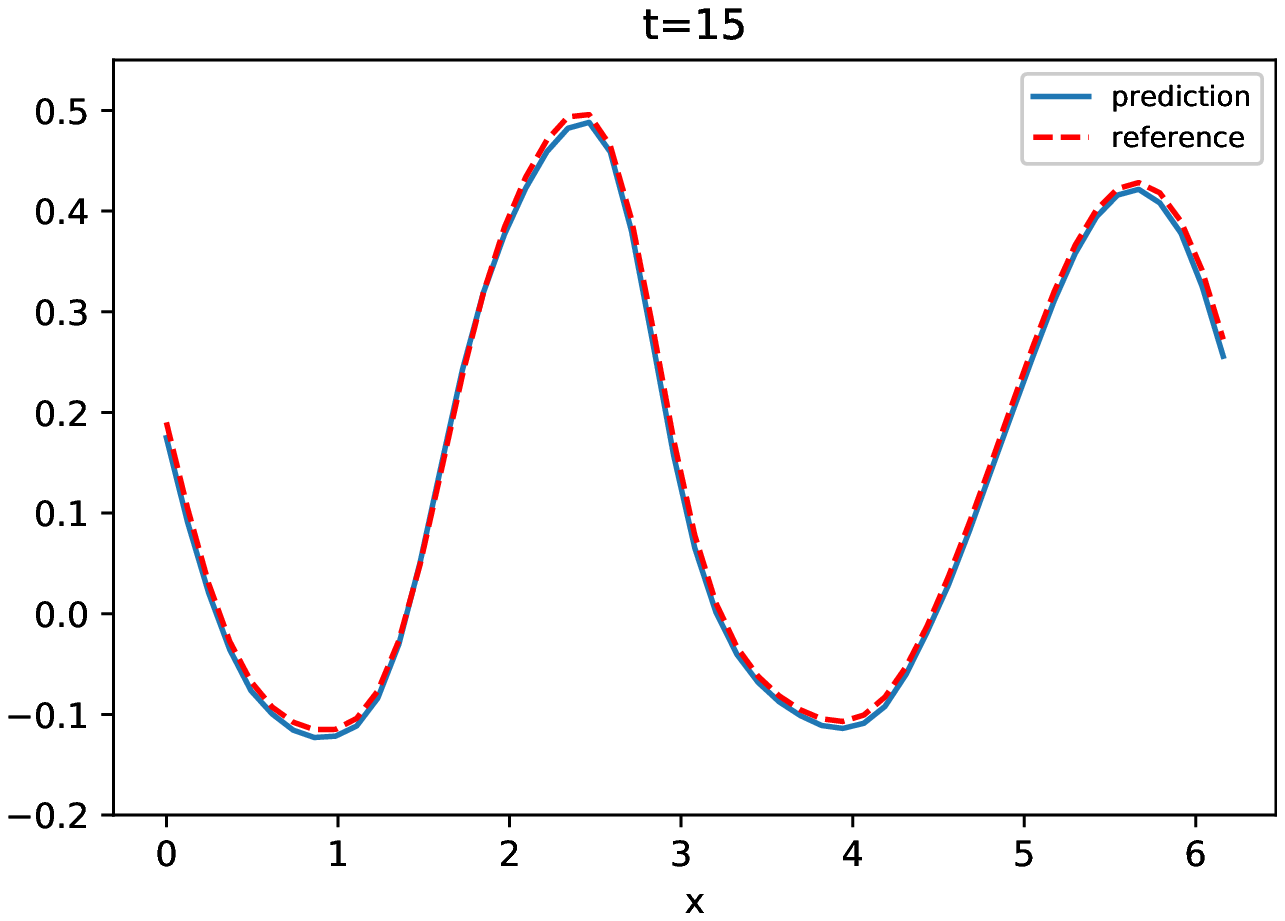}
		\includegraphics[width=0.49\textwidth]{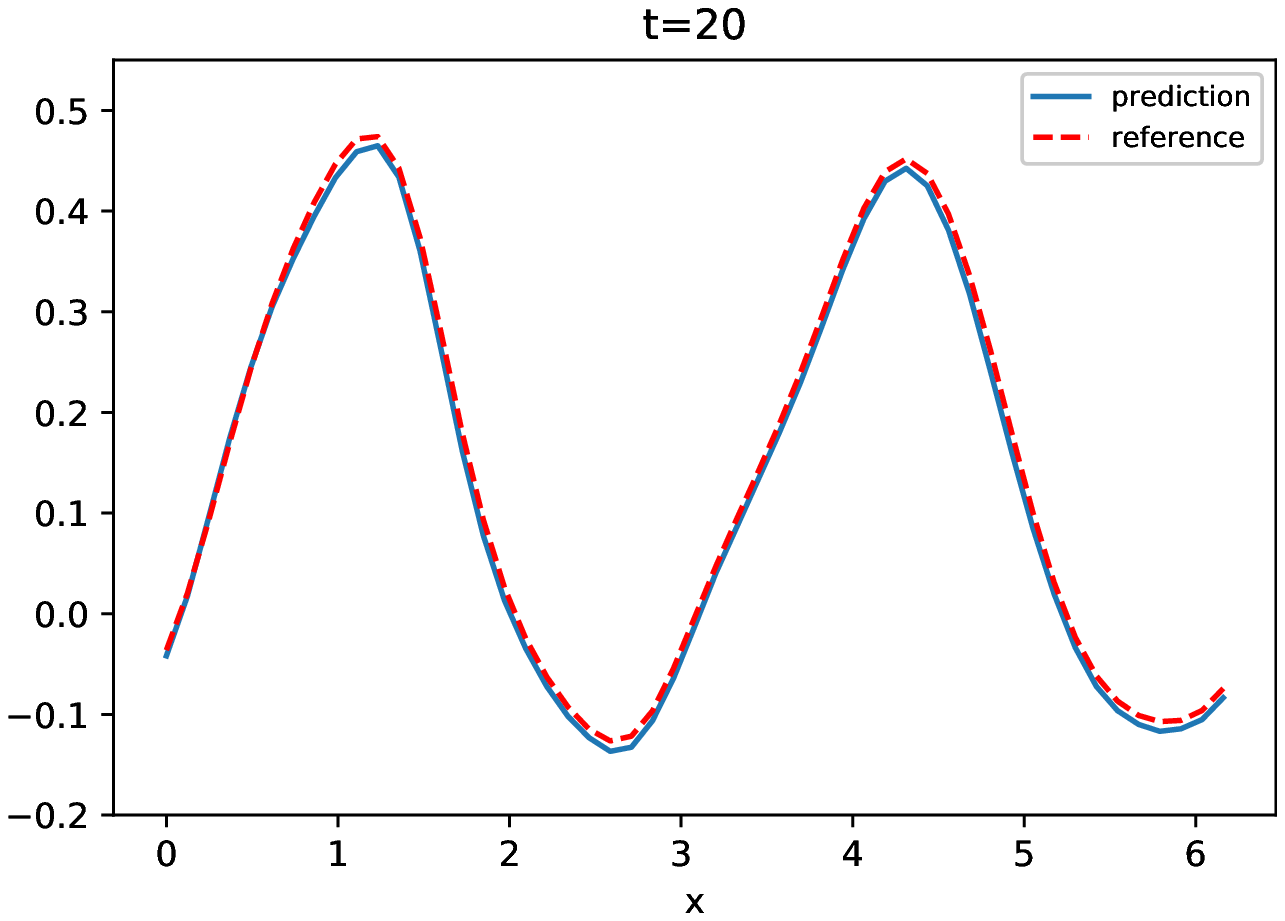}
		\caption{Advection-diffusion on uniform grids. Comparison between DNN model
                  prediction and reference solution.}
		\label{fig:ex1_adv-diff}
	\end{center}
\end{figure}

We first consider DNN modeling on uniform grids in physical space. In particular, we consider the nodes $x_i = \frac{2\pi}{N}i$, $i = 0,...,N-1$, where $N = 50$.
To validate the trained DNN model, we conduct numerical predictions with initial conditions not in the training data set.
Here, we present the results by an initial condition $u(x,0) = \exp(-\sin^2(x)) - 1/2$. In Fig. \ref{fig:ex1_adv-diff}, the DNN predictions are plotted at different time instances, along with the reference solutions (by solving the
true model using the same initial condition) for comparison. We observe that the DNN model produces accurate prediction results for up to $t = 20$, which is a much longer time span than the training data set (for which $t=0.2$).

\begin{figure}[htbp]
	\begin{center}
		\includegraphics[width=0.49\textwidth]{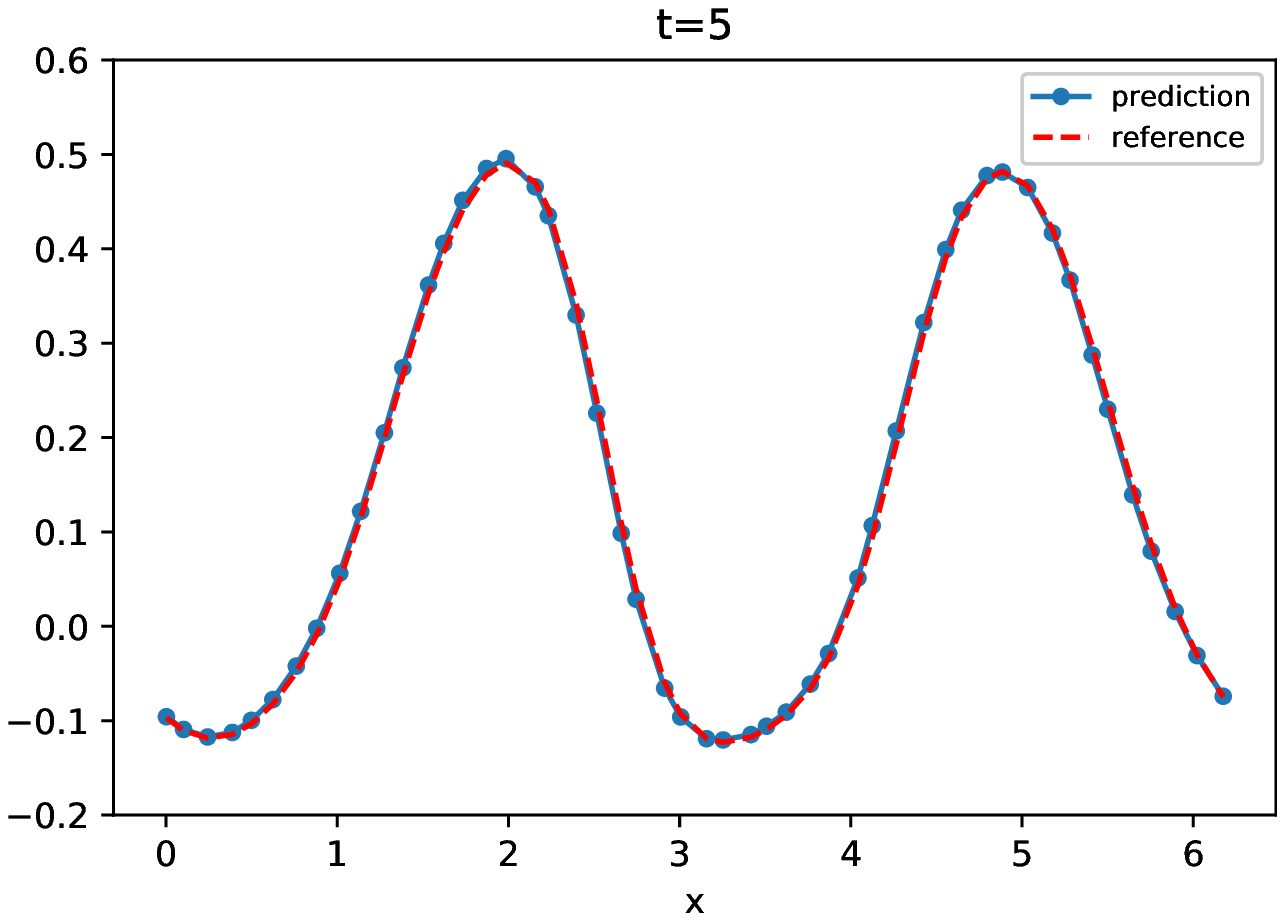}
		\includegraphics[width=0.49\textwidth]{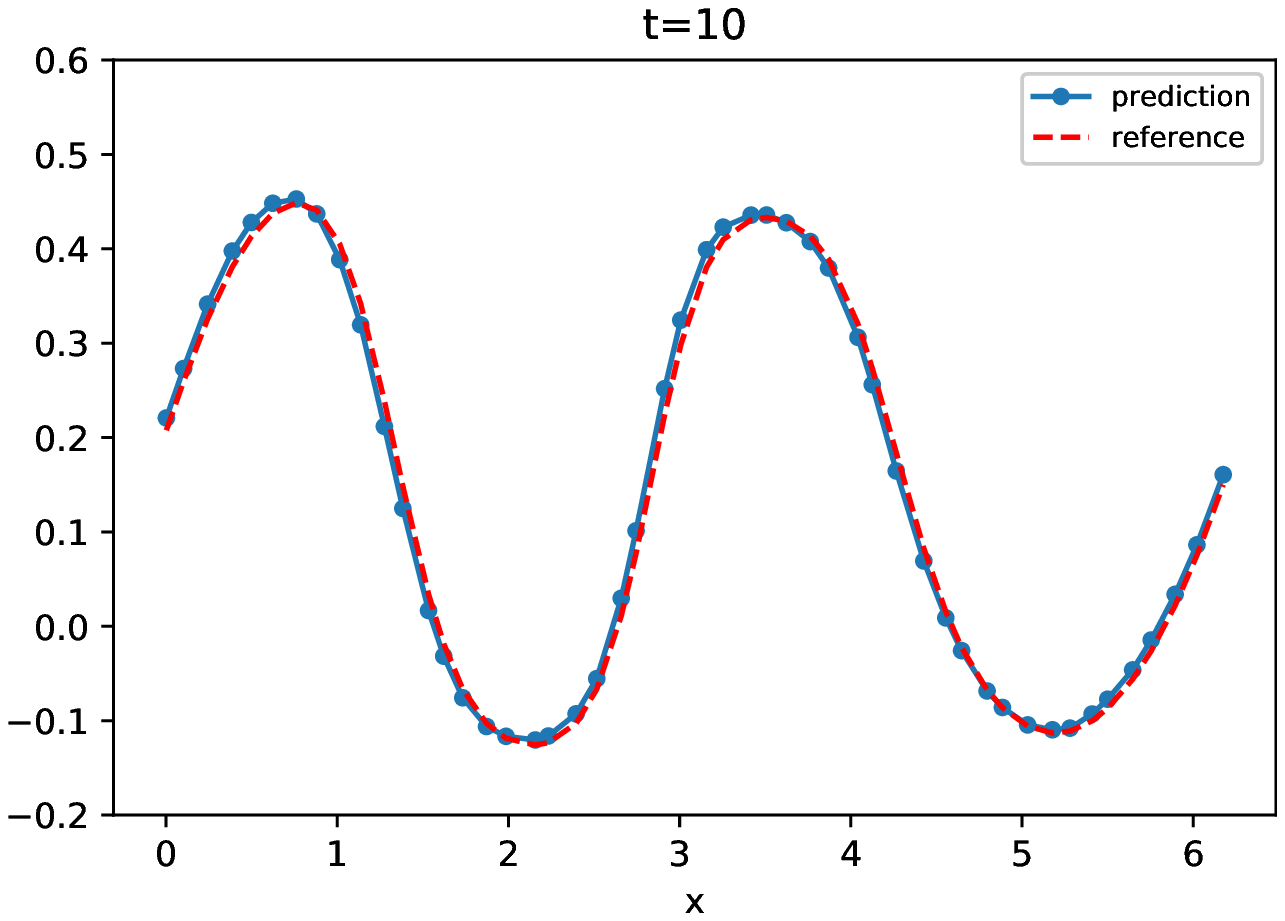}
		\includegraphics[width=0.49\textwidth]{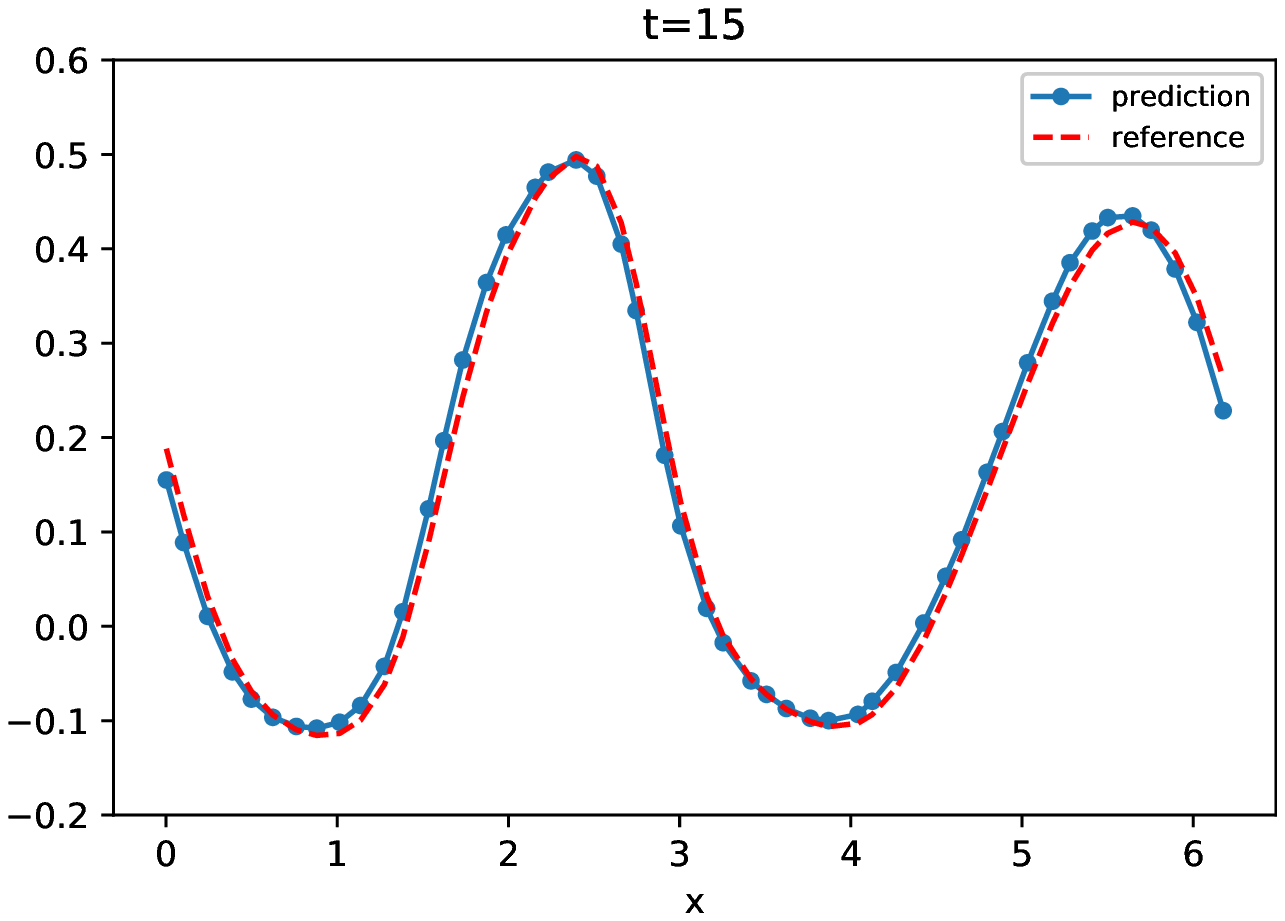}
		\includegraphics[width=0.49\textwidth]{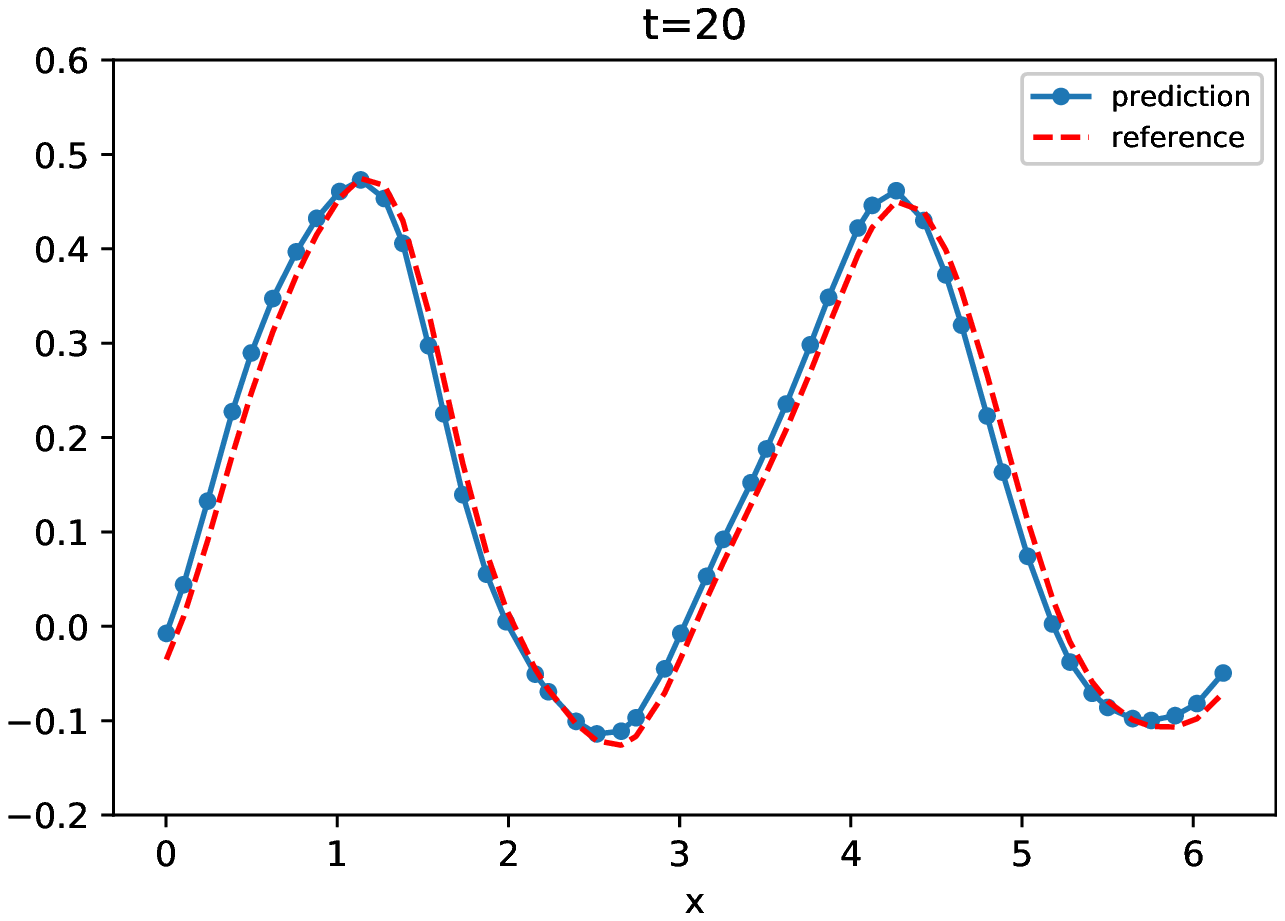}
		\caption{Advection-diffusion on unstructured grids. Comparison between DNN model
                  prediction and reference solution. The non-uniform grids are the solid dots,
                  whose indices are 
                  randomly permuted.}
		\label{fig:ex1_adv-diff_rand}
	\end{center}
\end{figure}

We then consider DNN modeling on an unstructured grid. Specifically, we randomly perturb the uniform grid from above
by up to $\pm 25\%$ of the interval size. We then randomly permute the
grid ordering. The resulting grids are therefore void of
any geometric information. We then train the DNN on this randomized
grid set. The DNN predictions using the initial condition $u(x,0) = \exp(-\sin^2(x)) - 1/2$
are shown in Fig. \ref{fig:ex1_adv-diff_rand}. We again observe good agreement with the reference solution for
time up to $t=20$. This demonstrates the flexibility of the proposed
DNN modeling, which can be completely structure-free.

\subsection{4th-order PDE}

We now consider a 4th-order PDE, 
\begin{equation} \label{laplacian}
 \frac{\partial u}{\partial t} + c\frac{\partial^4 u}{\partial x^4} = 0, 
\end{equation}
with $c = 1 \times 10^{-2}$ and $2\pi$-periodic boundary condition.

We consider a uniform grid with nodes given by $x_i = \frac{2\pi}{N}i, i = 0,...,N-1$, with $N = 50$.
The training data are generated using the Fourier series \eqref{Fourier}, with $N_c=7$ and uniformly distributed
random coefficients $a_0\sim [-1/2,1/2]$, and $a_n^{(i)}, b_n^{(i)} \sim U[-\frac{1}{n}, \frac{1}{n}]$, $1 \leq n \leq N_c$.
The time step is $\Delta t = 1 \times 10^{-2}$, and the length of the
training trajectories is $n_L=10$, as in \eqref{loss_n}.
The training data set consists of $10,000$ such trajectories.
The $\tanh$ activation function is used throughout, along with cyclic learning with maximum learning rate of $1\times10^{-3}$, minimum learning rate of $1\times10^{-4}$, and exponential decay rate $0.99994$. The DNN model is trained
for $2,000$ epochs with
batch size $50$.

\begin{figure}[htbp]
	\begin{center}
		\includegraphics[width=0.49\textwidth]{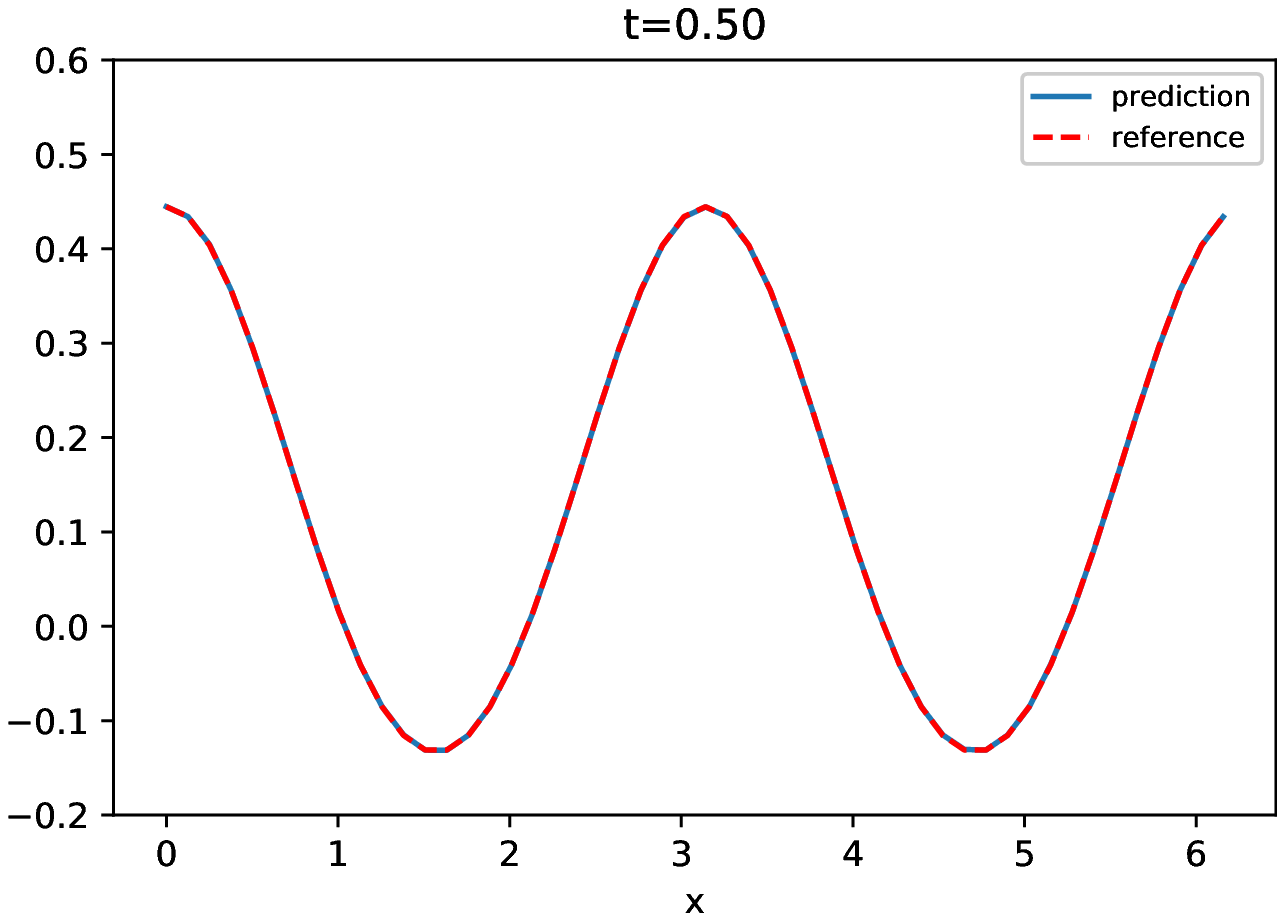}
		\includegraphics[width=0.49\textwidth]{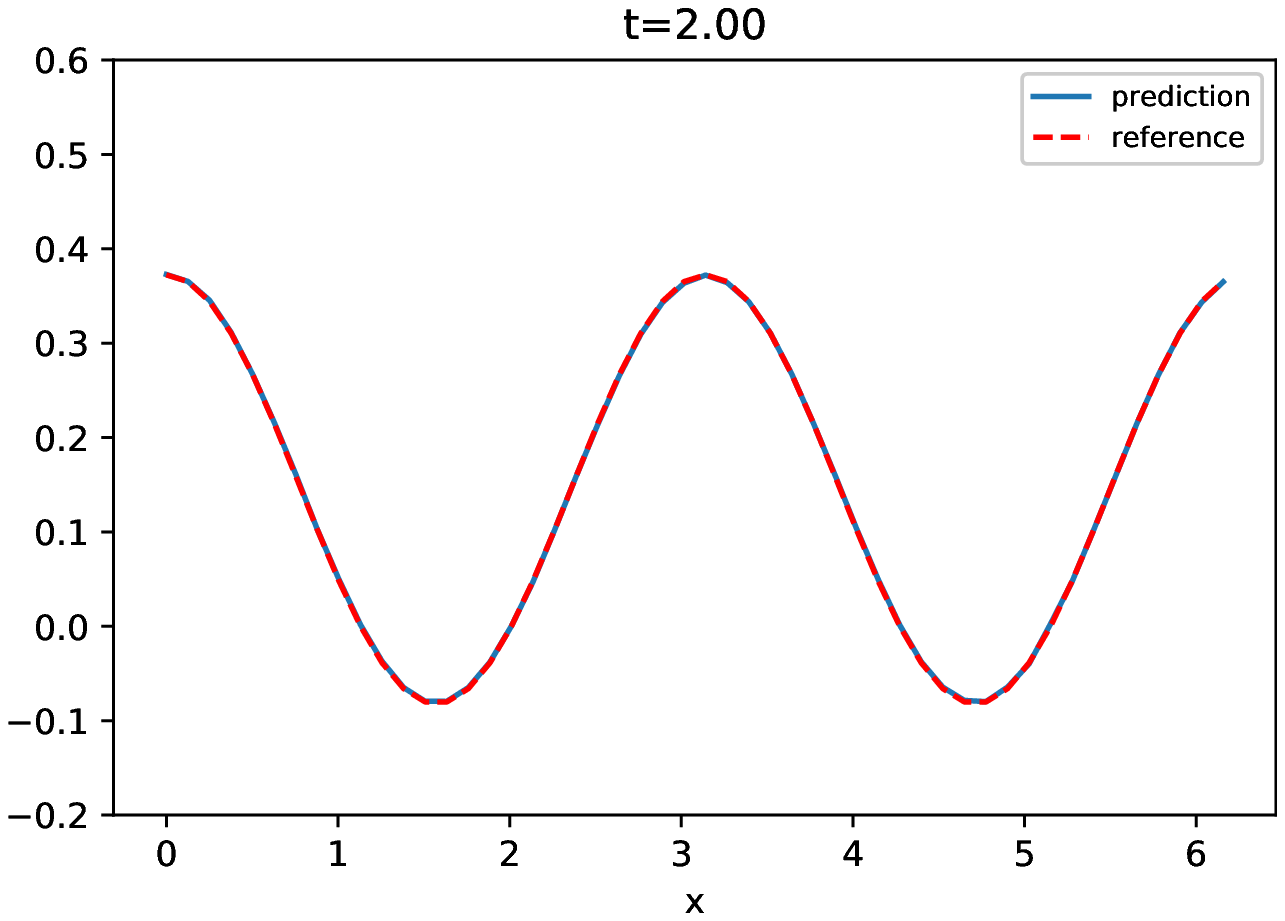}
		\includegraphics[width=0.49\textwidth]{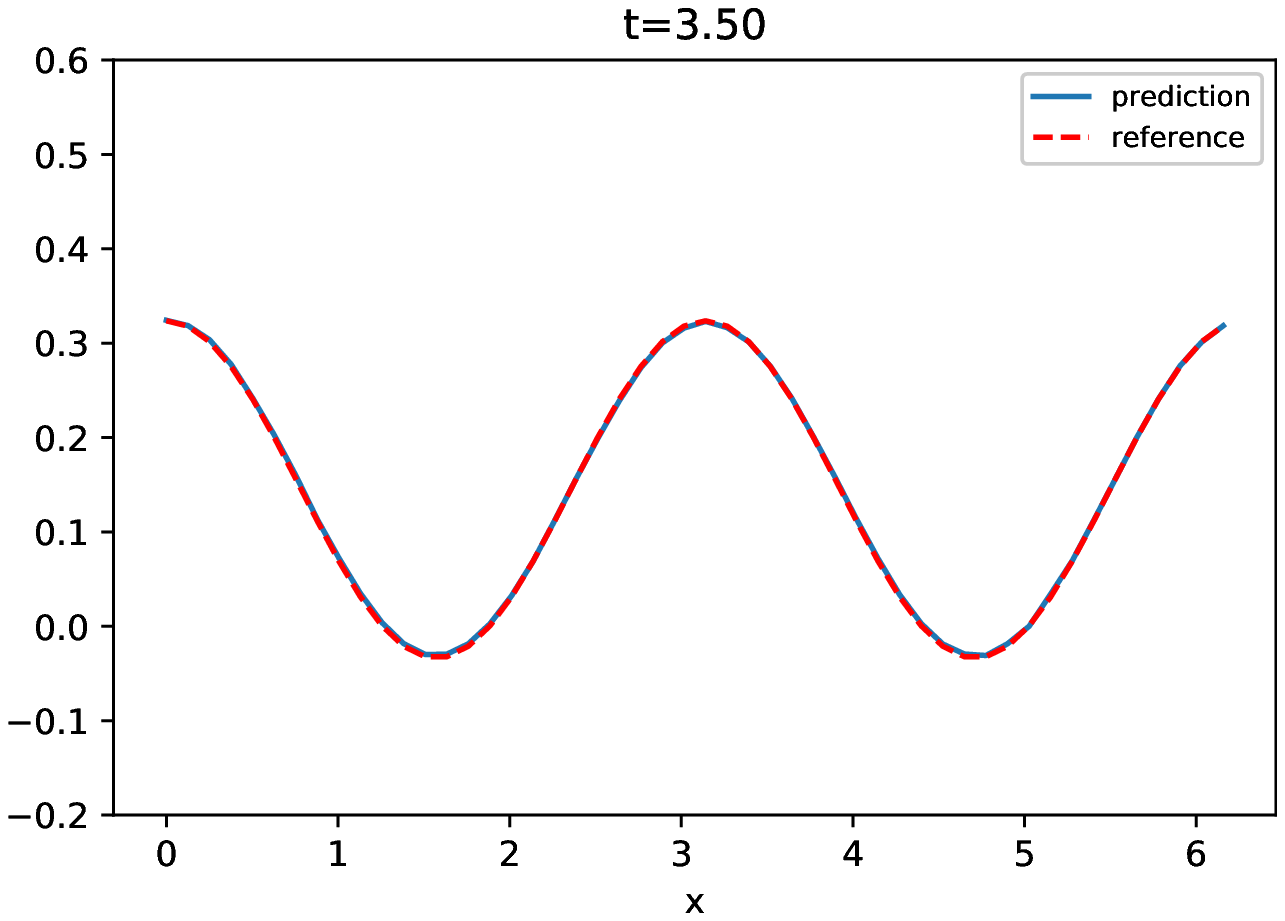}
		\includegraphics[width=0.49\textwidth]{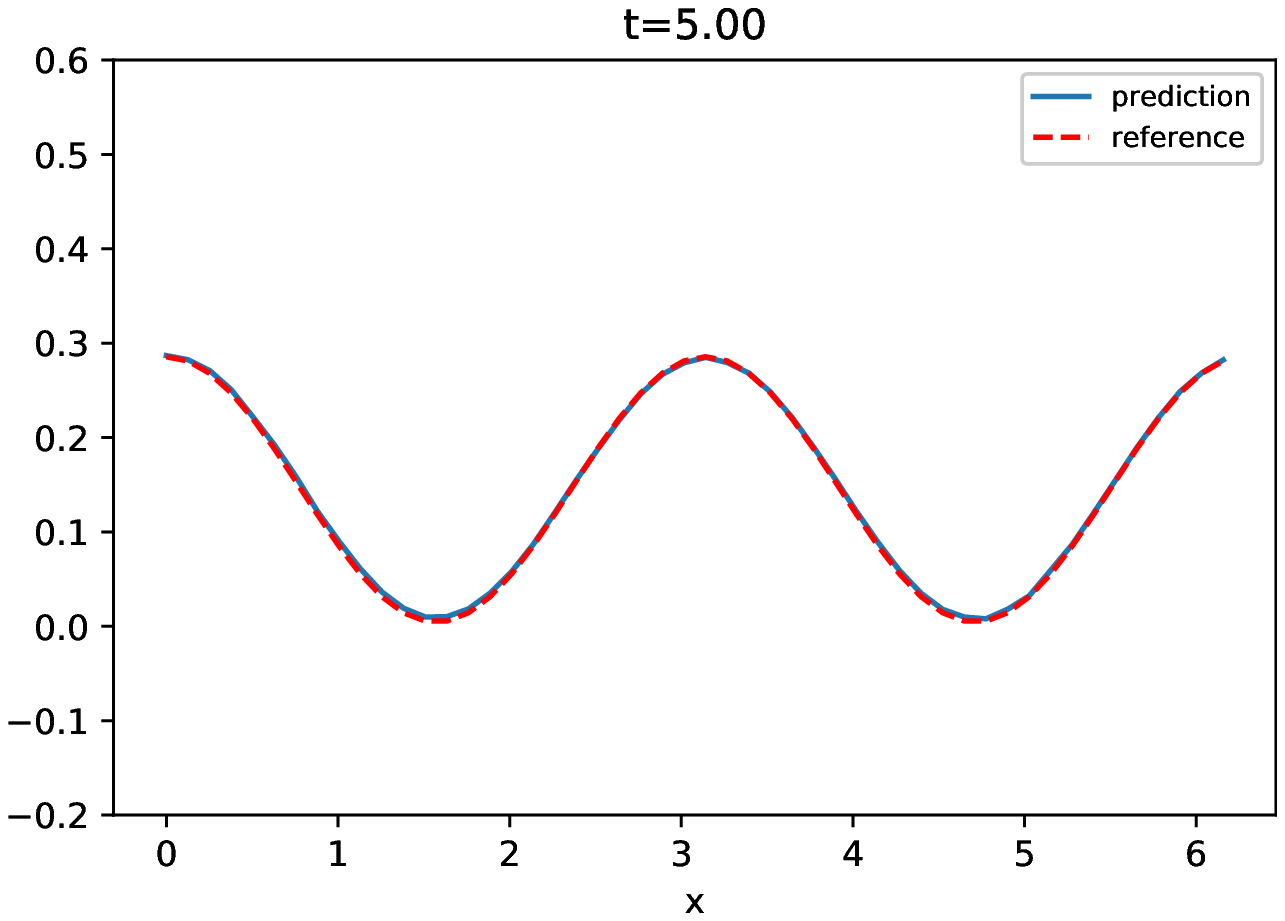}
		\caption{4th-order PDE: Comparison between
                  DNN model prediction and reference solution.}
		\label{fig:ex2_double-lap}
	\end{center}
\end{figure}

Validation results with the initial condition $u(x,0) = \exp(-\sin^2(x)) - 1/2$ are shown in Fig. \ref{fig:ex2_double-lap}.
We observe excellent agreement between the DNN model predictions and the reference solutions over time.

\subsection{Differential-Integral Equation}

As discussed in Section \ref{sec:others}, the proposed DNN structure can be used to model other
types of equations involving operators such as integrals, see \eqref{integral}. Here, we
consider a differential-integral equation,
\begin{equation}
 \frac{\partial u}{\partial t} + \frac{\partial u}{\partial x} = \frac{1}{2\pi} \int_0^{2\pi} u(y,t) dy - u,
\end{equation}
with $2\pi$-periodic boundary condition.

The training data are generated on a uniform grid with nodes given by
$x_i = \frac{2\pi}{N}i$, $i = 0,...,N-1$ with $N = 50$. A total number of $10,000$ trajectories with time
step $\Delta t = 5 \times 10^{-4}$ are stored over $n_L=10$ steps \eqref{loss_n}. Note that the training trajectory data are
of very short temporal length, only up to $t=0.005$.  The trajectories are generated by
solving the true equation with the Fourier initial condition \eqref{Fourier} with $a_0=0$, $a_n, b_n \sim U[-1,1]$, and $N_c\sim U\{0,\dots, 10\}$.

The hyperbolic tangent function $\tanh$ is used as the activation
function, along with
cyclic learning with maximum learning rate of $1\times10^{-3}$,
minimum learning rate of $1\times10^{-4}$, and exponential decay rate of
$0.99994$. The DNN model was trained for  
$30,000$ epochs with batch size 50. 

\begin{figure}[htbp]
	\begin{center}
		\includegraphics[width=0.49\textwidth]{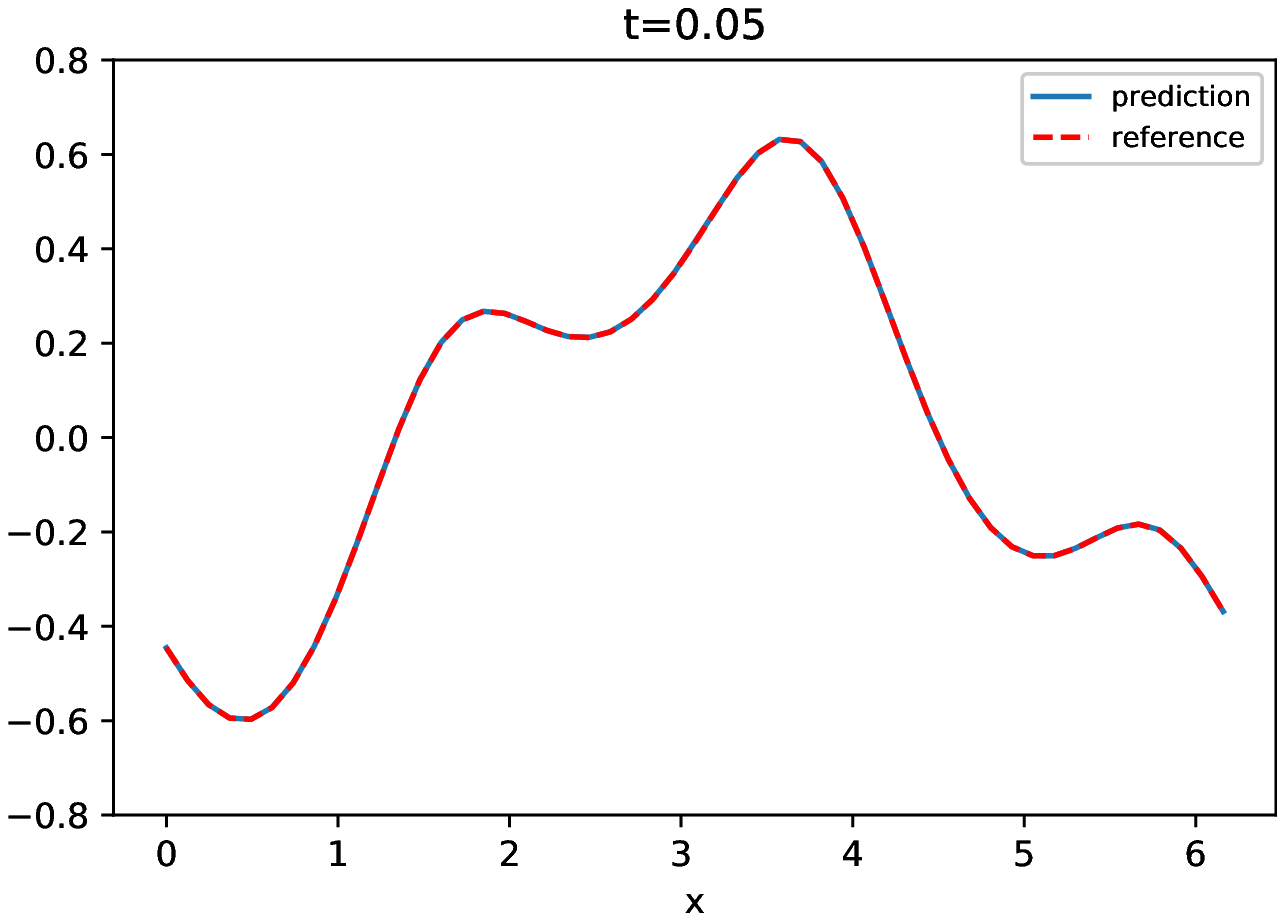}
		\includegraphics[width=0.49\textwidth]{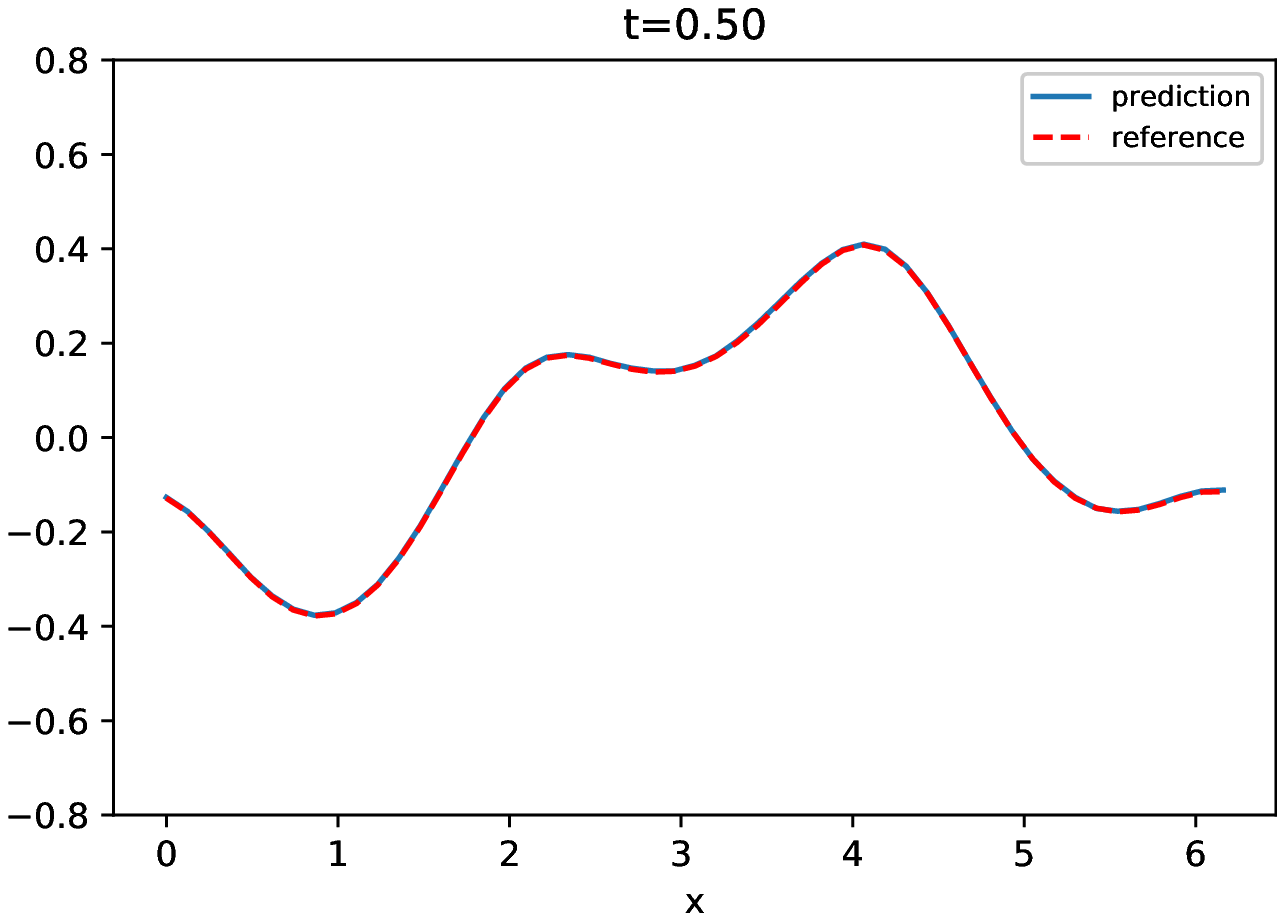}
		\includegraphics[width=0.49\textwidth]{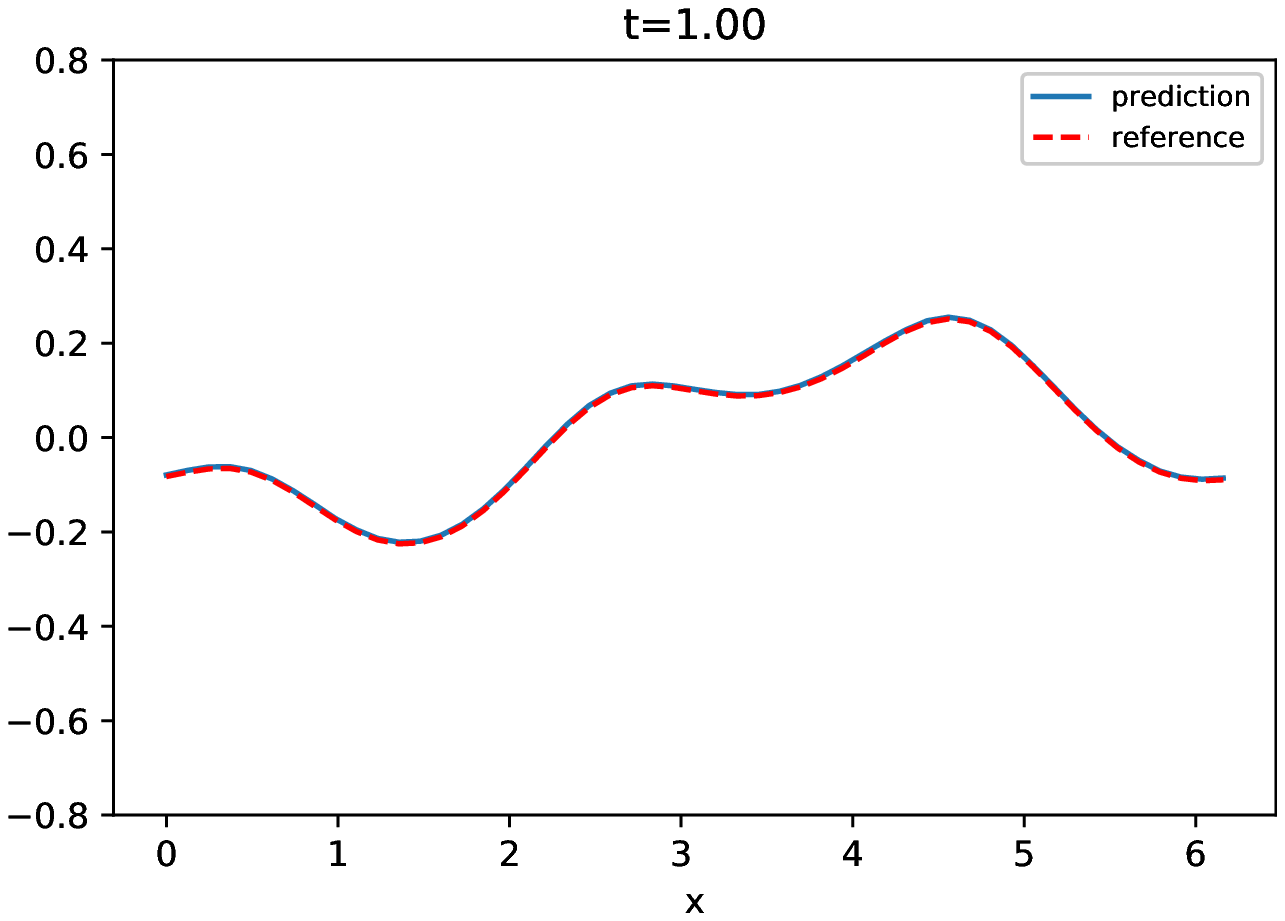}
		\includegraphics[width=0.49\textwidth]{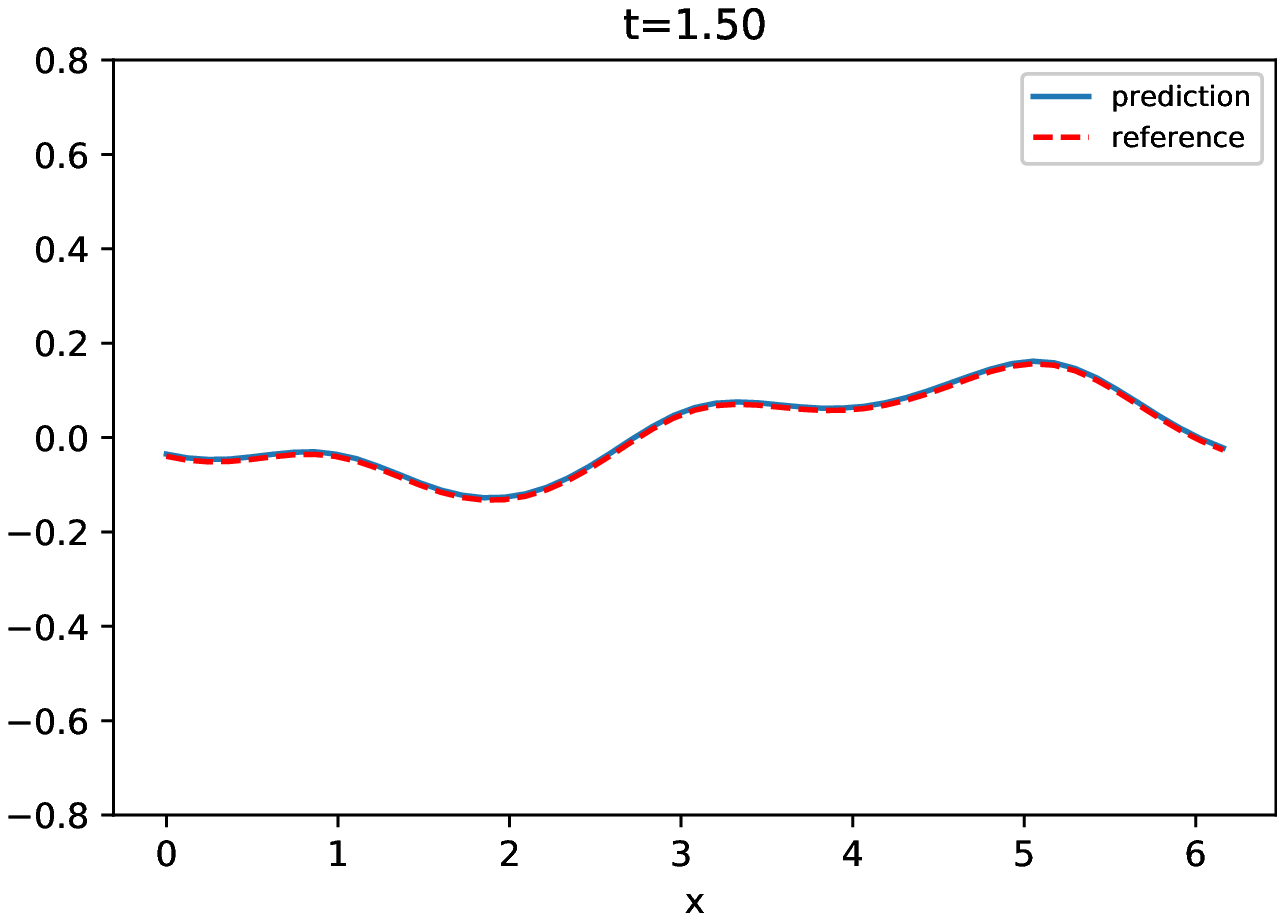}
		\caption{Differential-integral
                  equation: Comparison between DNN model prediction and reference solution.}
		\label{fig:ex3_diff-integ}
	\end{center}
\end{figure}

In Fig. \ref{fig:ex3_diff-integ}, the validation results obtained using the initial condition
$$u(x,0) = \exp\left(-\frac{1}{5}\sin(3x)\right) - \frac{1}{2}\cos(x) - 1
$$ are shown. We observe excellent agreement with the reference solution, obtained by solving the true equation.
The results are up to $t=1.5$, which is significantly longer than the temporal length of the training data (for which $t=0.005$).

\subsection{Burger's Equation}

In this section we consider Burger's equation, in both viscous and inviscid form. The latter case poses significant
numerical challenges, as shocks may develop in finite time.

\subsubsection{Viscous Burger's Equation}

We first consider viscous Burgers equation
\begin{equation} \label{burger1}
 \frac{\partial u}{\partial t}  + u \frac{\partial u}{\partial x}  = \nu \frac{\partial^2 u}{\partial x^2}, 
\end{equation}
with $2\pi$-periodic boundary condition and viscosity $\nu = 0.1$.

Similar to the other cases, the training data are generated by solving the true equation
\eqref{burger1} on a uniform grid with nodes given by $x_i = -\pi + \frac{2\pi}{N}i$, $i = 0,...,N-1$ where $N = 50$. The
Fourier series \eqref{Fourier} is used to generate $10,000$ randomized trajectory data using coefficients $a_0\sim U[-1/2, 1/2]$
and $a_n, b_n \sim U[-\frac{1}{n}, \frac{1}{n}]$ for $1 \leq n \leq
N_c$, and $N_c=10$. The trajectories are of length $n_L=10$ and with 
time step $\Delta t = 0.01$, for the multi-step loss computation
\eqref{loss_n}.

Cyclic learning is used, with maximum learning rate $1\times10^{-3}$,
minimum learning rate $1\times10^{-4}$, and an exponential decay rate
$0.99994$, along with $\tanh$ as the activation function. 
The DNN was trained for $20,000$ epochs with batch size 50.

\begin{figure}[htbp]
	\begin{center}
		\includegraphics[width=0.49\textwidth]{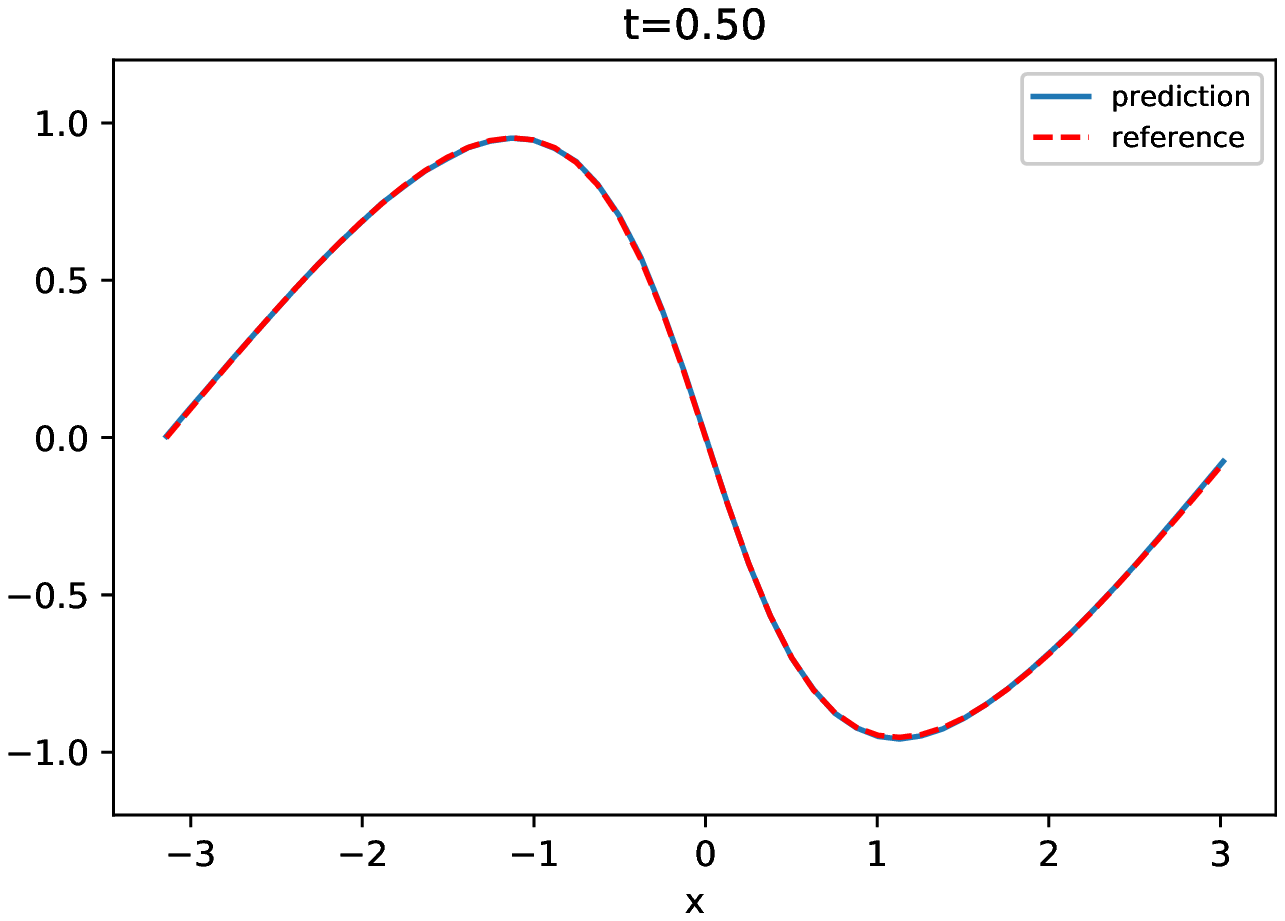}
		\includegraphics[width=0.49\textwidth]{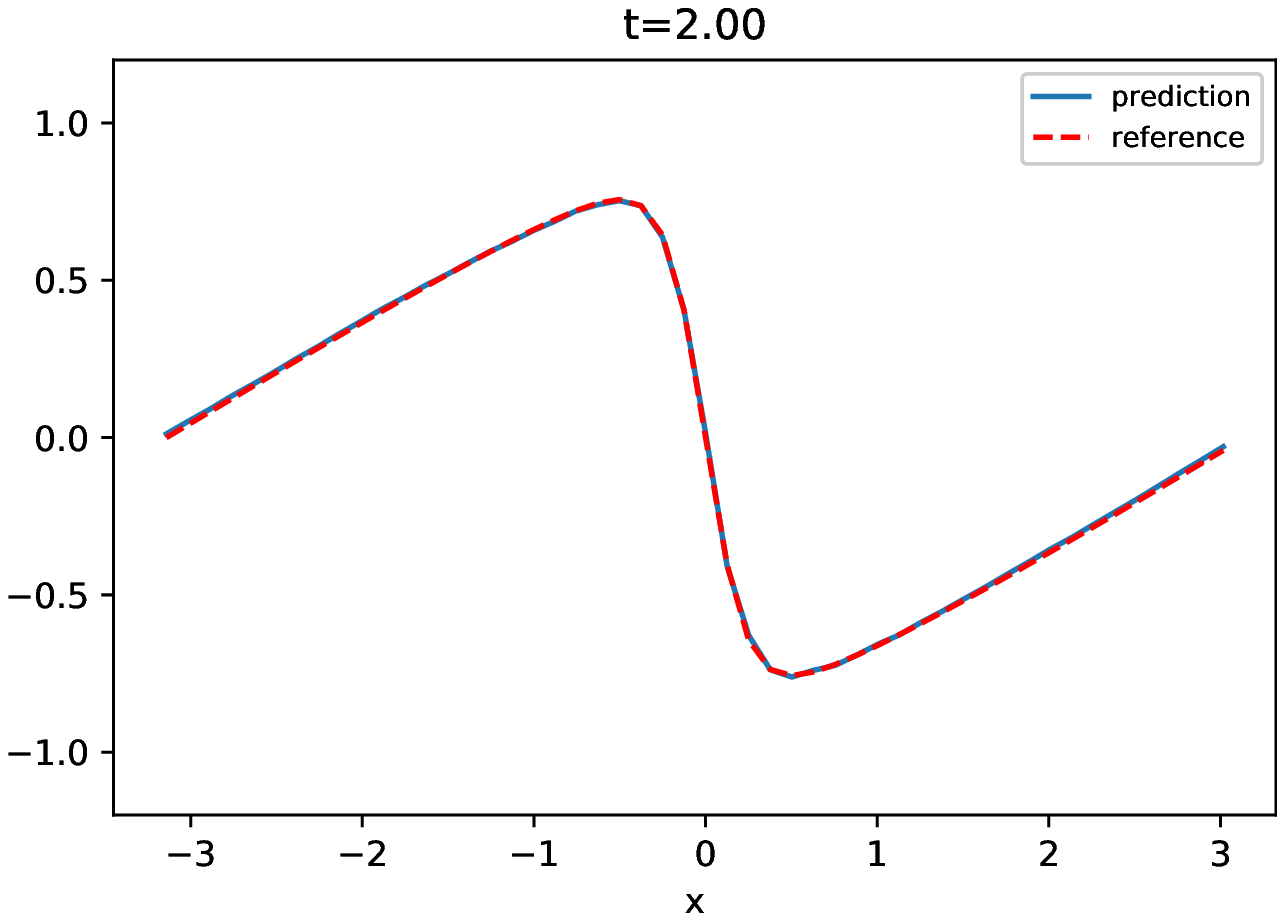}
		\includegraphics[width=0.49\textwidth]{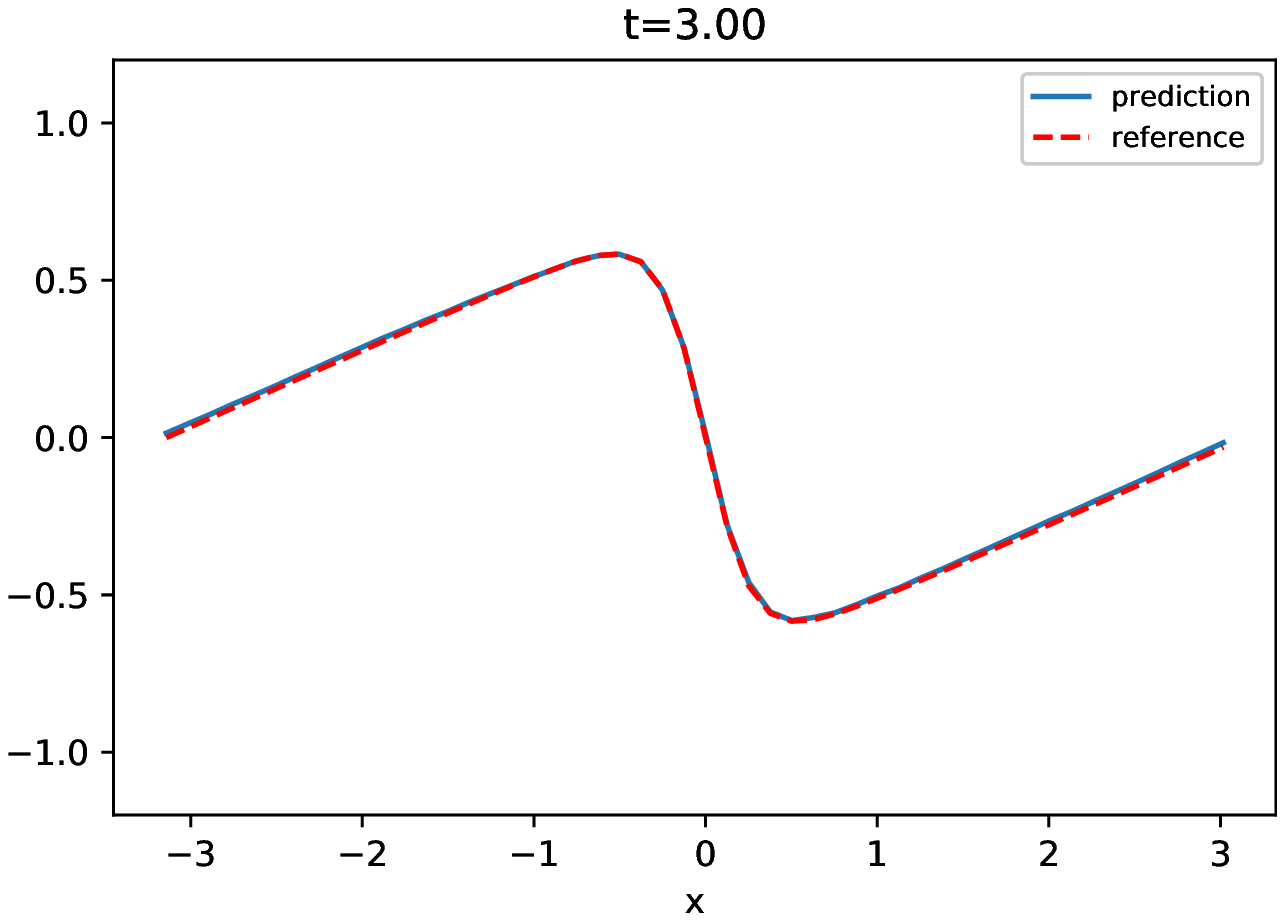}
		\includegraphics[width=0.49\textwidth]{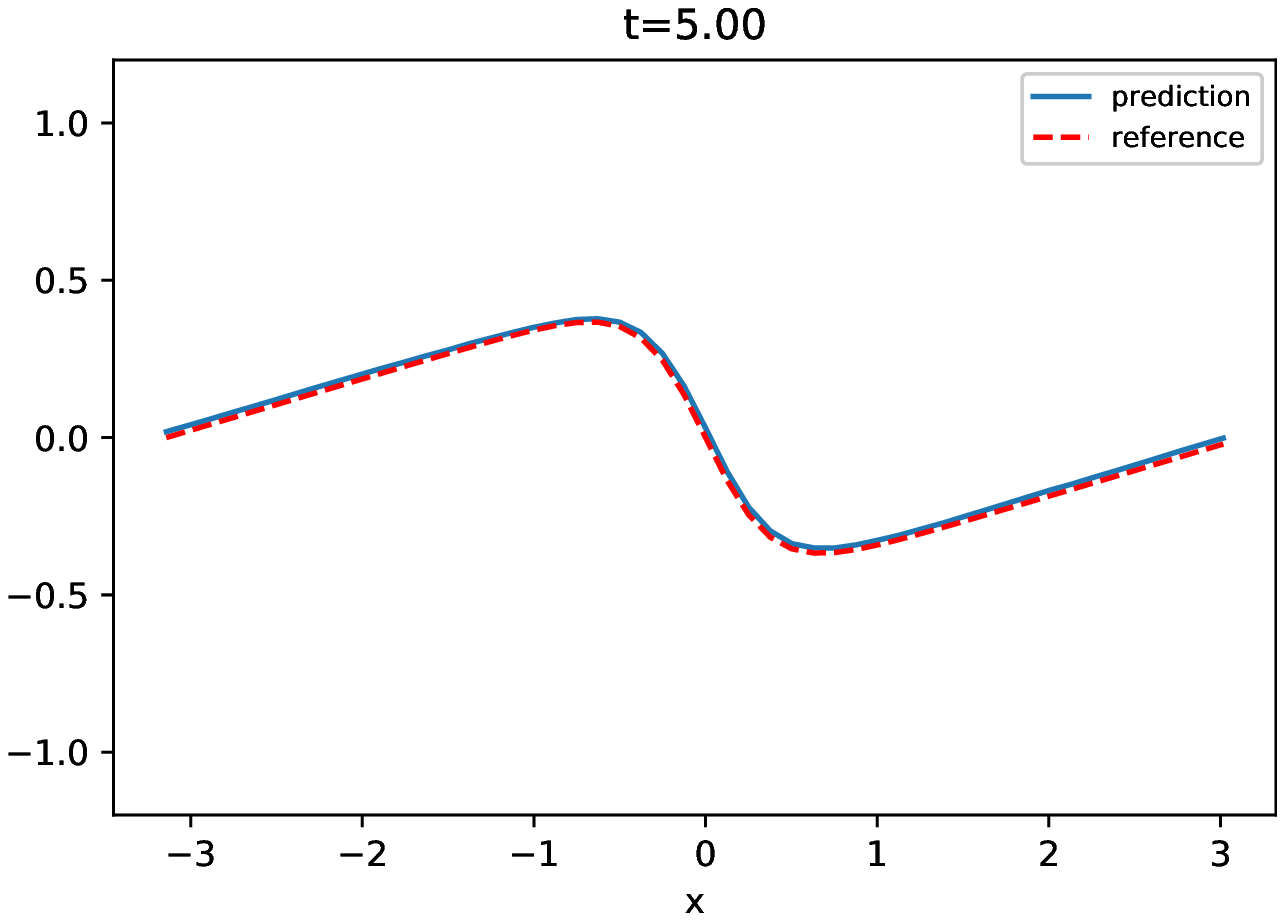}
		\caption{Viscous Burger's
                  equation with viscosity $\nu=0.1$.  Comparison of DNN model prediction and reference solution.}
		\label{fig:ex4_visBurg_v1e-01}
	\end{center}
\end{figure}

In Fig. \ref{fig:ex4_visBurg_v1e-01}, we present the DNN prediction under initial condition $u(x,0) = \sin(x)$, along
with the reference solution. We observe that the DNN generates an accurate prediction up to $t=5$.
Note that the solution starts to develop a sharper slope early on, before the viscous term takes effect and diffuses it away.
Without the viscous term, a shock will develop, as shown in the next example.

\subsubsection{Inviscid Burger's Equation}
      
We now consider inviscid Burgers equation with $2\pi$-periodic boundary condition,
\begin{equation}
 \frac{\partial u}{\partial t}  + \frac{\partial}{\partial x}\left(\frac{1}{2} u^2\right)  = 0.
\end{equation}

The training data set for this example contains $20,000$ trajectory data. The first $10,000$ are generated in the same manner as in the viscous Burger's case
in the previous section. In addition, we include $10,000$ discontinuous solution data, which are generated from
randomized piecewise constant initial data
\begin{equation}
	u_0(x) = 
	\begin{cases}
		u_1, &  x_\ell \leq x \leq x_r\\
		u_2, & \textrm{ elsewhere}
	\end{cases},       
      \end{equation}
      where $u_1, u_2 \sim U[-1,1]$ and $x_\ell, x_r \sim U[-\pi, \pi]$.
      For each of these piecewise constant initial conditions, we generated $n_L=10$ steps of training trajectory
      data with time step $\Delta t = 0.01$, by numerically solving
      the true equation using 9th-order WENO method in space and
      4th-order Runge-Kutta method in time. 

\begin{figure}[htbp]
	\begin{center}
		\includegraphics[width=0.49\textwidth]{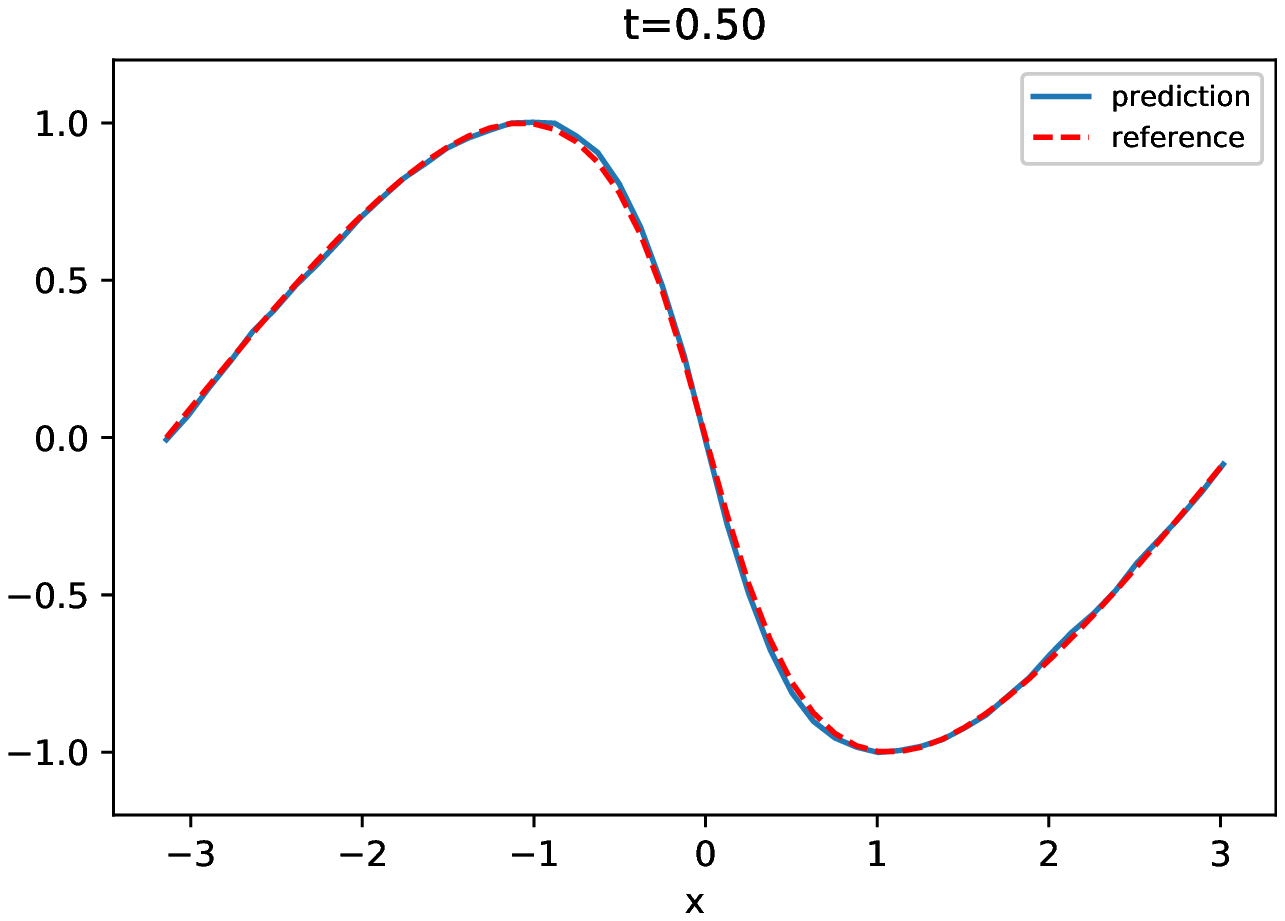}
		\includegraphics[width=0.49\textwidth]{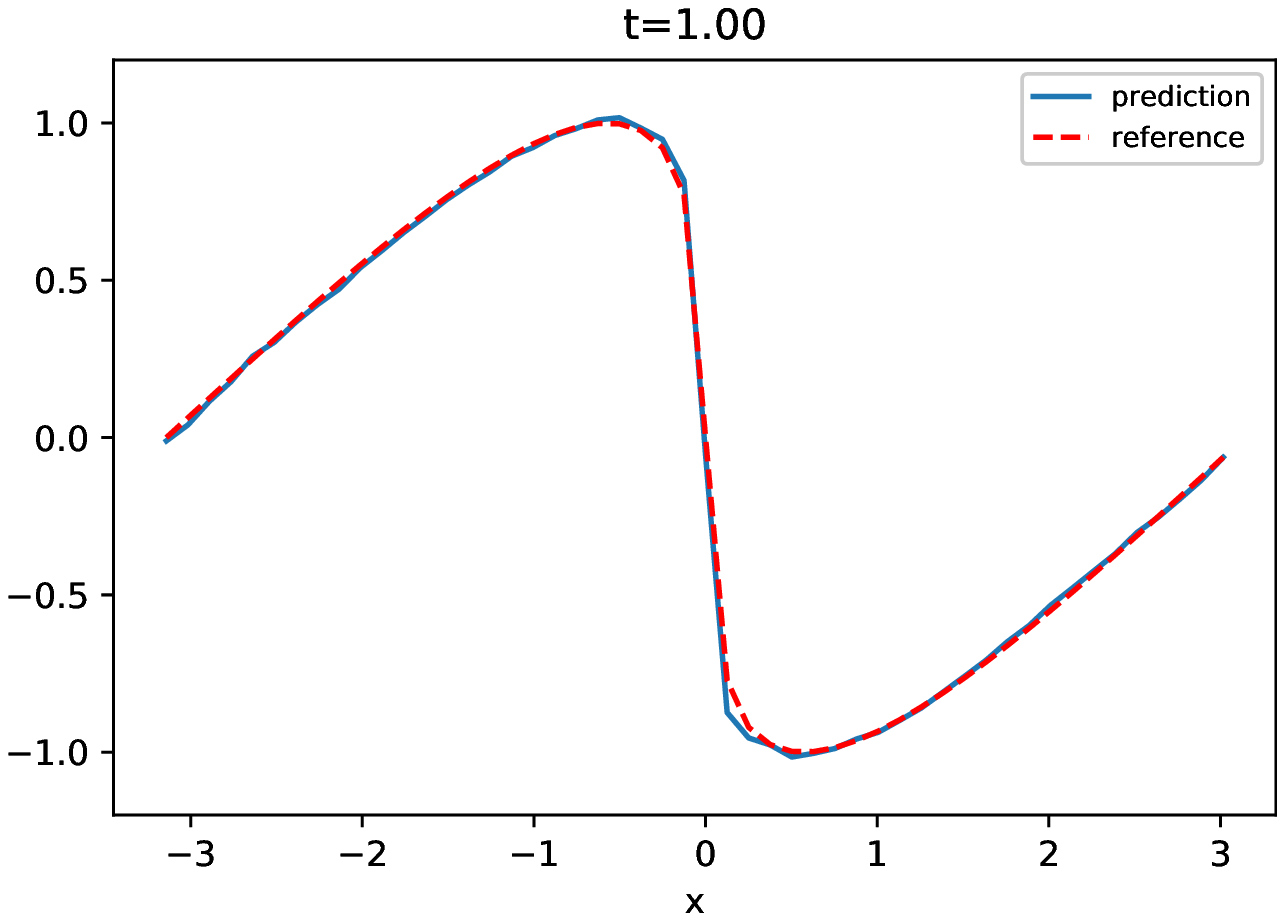}
		\includegraphics[width=0.49\textwidth]{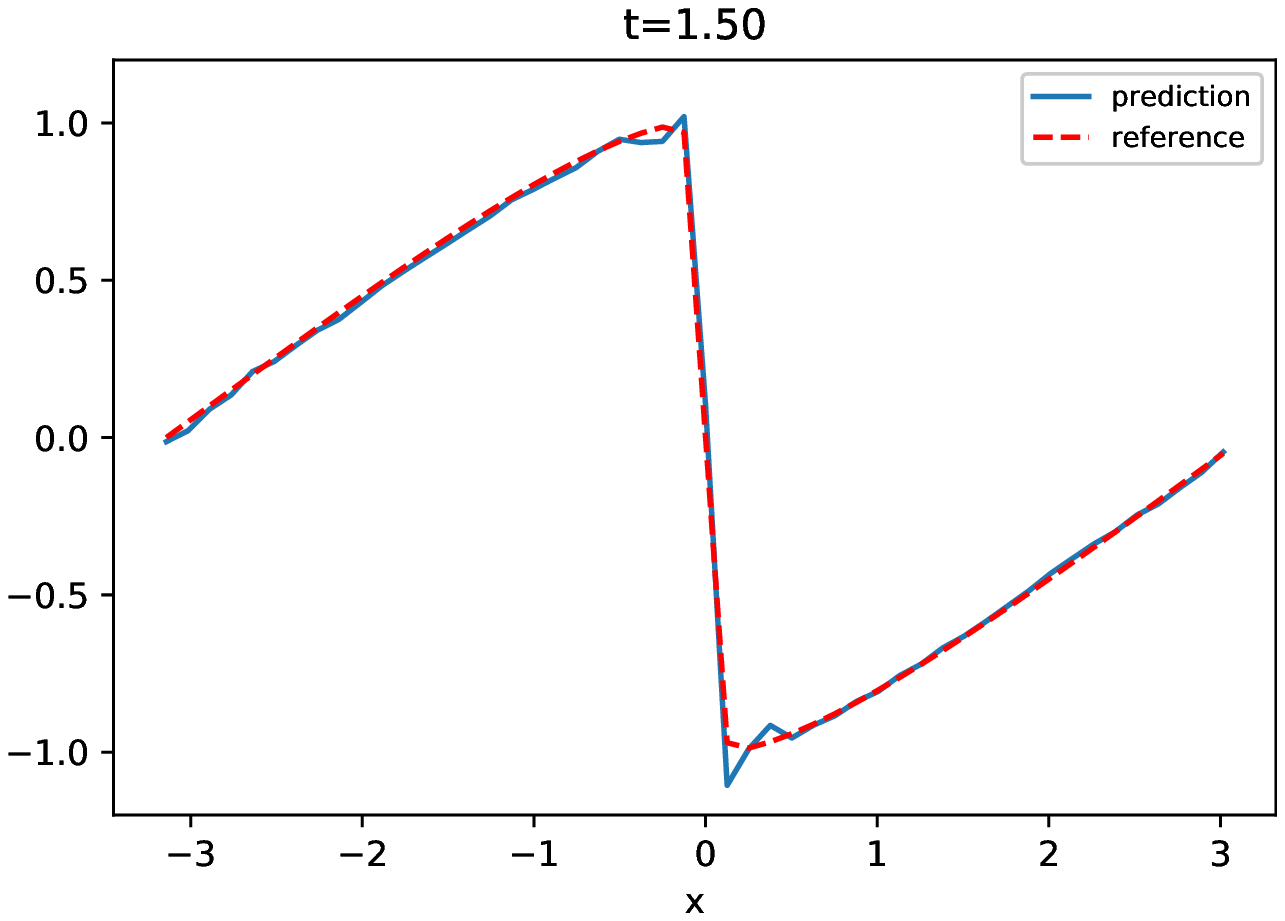}
		\includegraphics[width=0.49\textwidth]{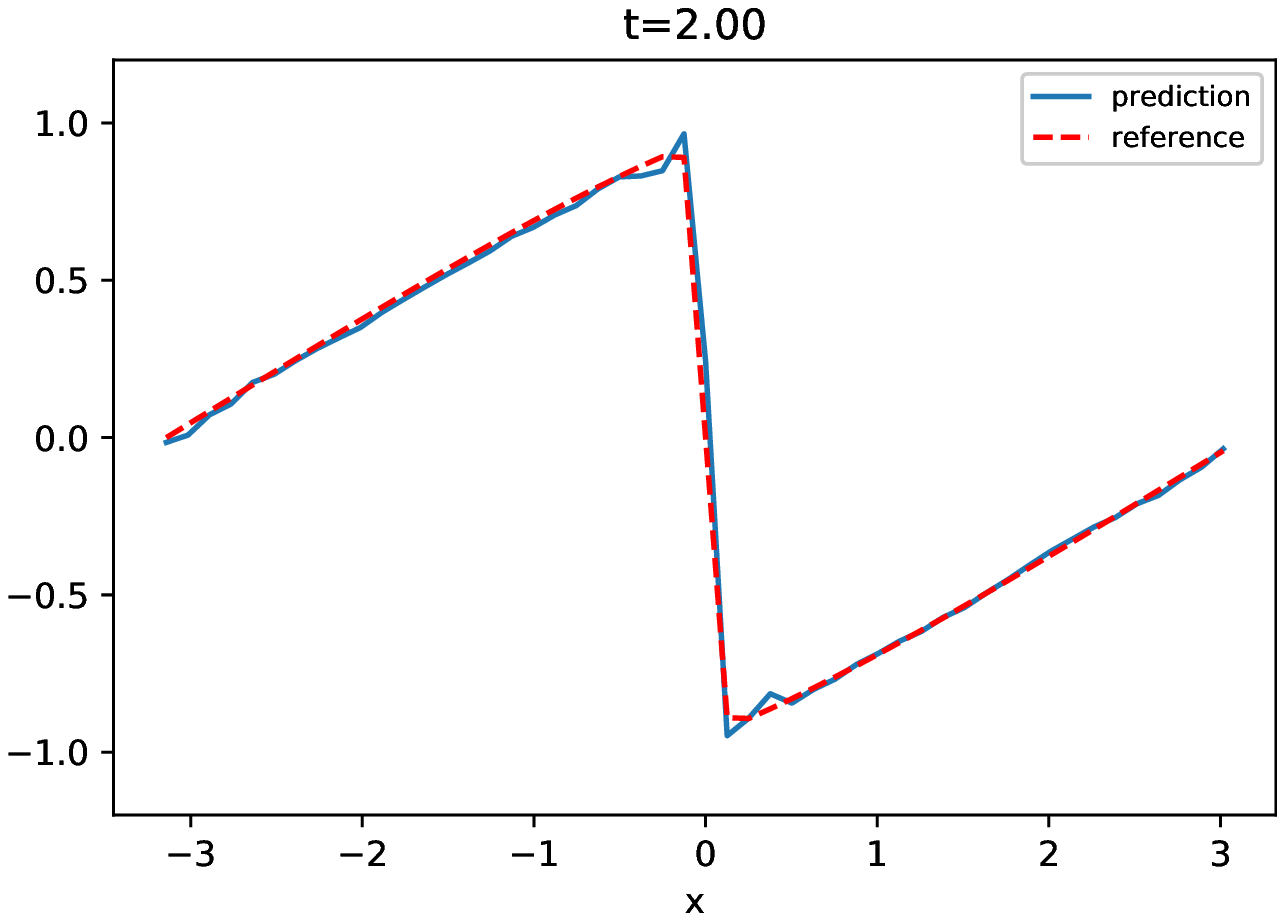}
		\caption{Inviscid Burger's
                  equation. Comparison of DNN model prediction and
                  reference solution. Left to right, up to
                  bottom. Note in this case shock develops at $t=2$.}
		\label{fig:ex4_invisBurg}
	\end{center}
\end{figure}

      The DNN structure and its training are set in the same manner as in the viscous Burger's example in the previous
      section. The prediction results generated by the trained DNN model using the initial condition $u(x,0) = \sin(x)$ are shown
      in Fig. \ref{fig:ex4_invisBurg}. In this case, the solution develops a shock
      discontinuity at time $t=2$, which the DNN model correctly predicts. 
      In particular, the predicted solution becomes
      sharper as time marches. At $t=1.5$, when the solution becomes very sharp near $x=0$, the DNN prediction
      exhibits small oscillations. This is a common feature of numerical solutions for sharp transition layers.
      At $t=2$, the shock is fully developed (as shown by the reference solution). The DNN prediction accurately
      captures the shock location and solution profile. Again, small numerical oscillations are visible near the shock
      location, as is common in most numerical solutions.

\subsection{PDE System}

We now discuss the proposed DNN learning for systems of PDEs. For
benchmarking purposes we consider a linear wave system
\begin{align}\label{eq:vecsystem}
\mathbf{u}_t = \A\mathbf{u}_x, 
\end{align}
where
\begin{align*}
\mathbf{u} = \begin{bmatrix} u_1(x,t) \\ u_2(x,t)\end{bmatrix} \quad\text{and}\quad \A=\begin{bmatrix} 0 & 1 \\ 1 & 0 \end{bmatrix},
\end{align*}
with $2\pi$-period boundary condition. (Note that even though its
form is simple, linear wave equations like this are exceptionally
difficult to solve accurately using numerical methods.)

The training data are collected over a uniform grid with $N=50$ points
and $n_L=10$ steps with a time step $\Delta t = 0.01$. They are the exact solution of the
true system under randomized initial conditions in the form of the
Fourier series \eqref{Fourier}. In particular, the initial conditions for
both components $u_1$ and $u_2$ are generated by sampling
\eqref{Fourier} with $a_0\sim U[-\frac{1}{2},\frac{1}{2}]$, and
$a_n ,b_n\sim U\left[-\frac{1}{n} ,\frac{1}{n}\right]$ for $1\le n \le N_c$ and $N_c=10$. A
total of $10,000$ trajectories are generated for the
training data set.

The DNN structure consists of an input layer with $100$ neurons, to
account for the two components $u_1$ and $u_2$, each of which has $50$
nodes.
The disassembly block has dimension $100\times 1 \times 5$ and the
assembly layer has dimension $1\times1\times 5$.
The DNN model is trained for $10,000$ epochs
with a constant learning rate of $10^{-3}$.

\begin{figure}[htbp]
	\begin{center}
		\includegraphics[width=.96\textwidth]{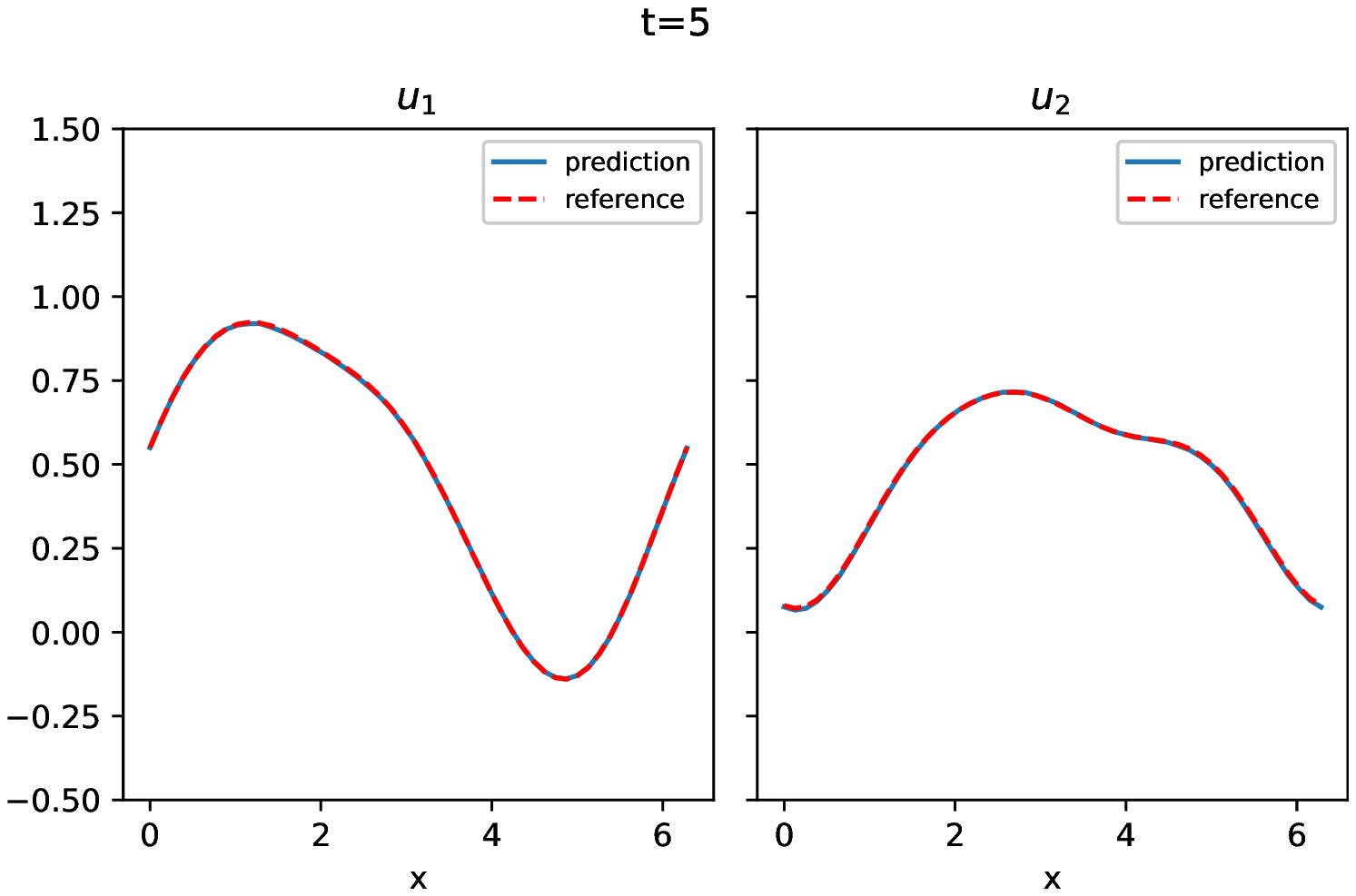}
		\includegraphics[width=.96\textwidth]{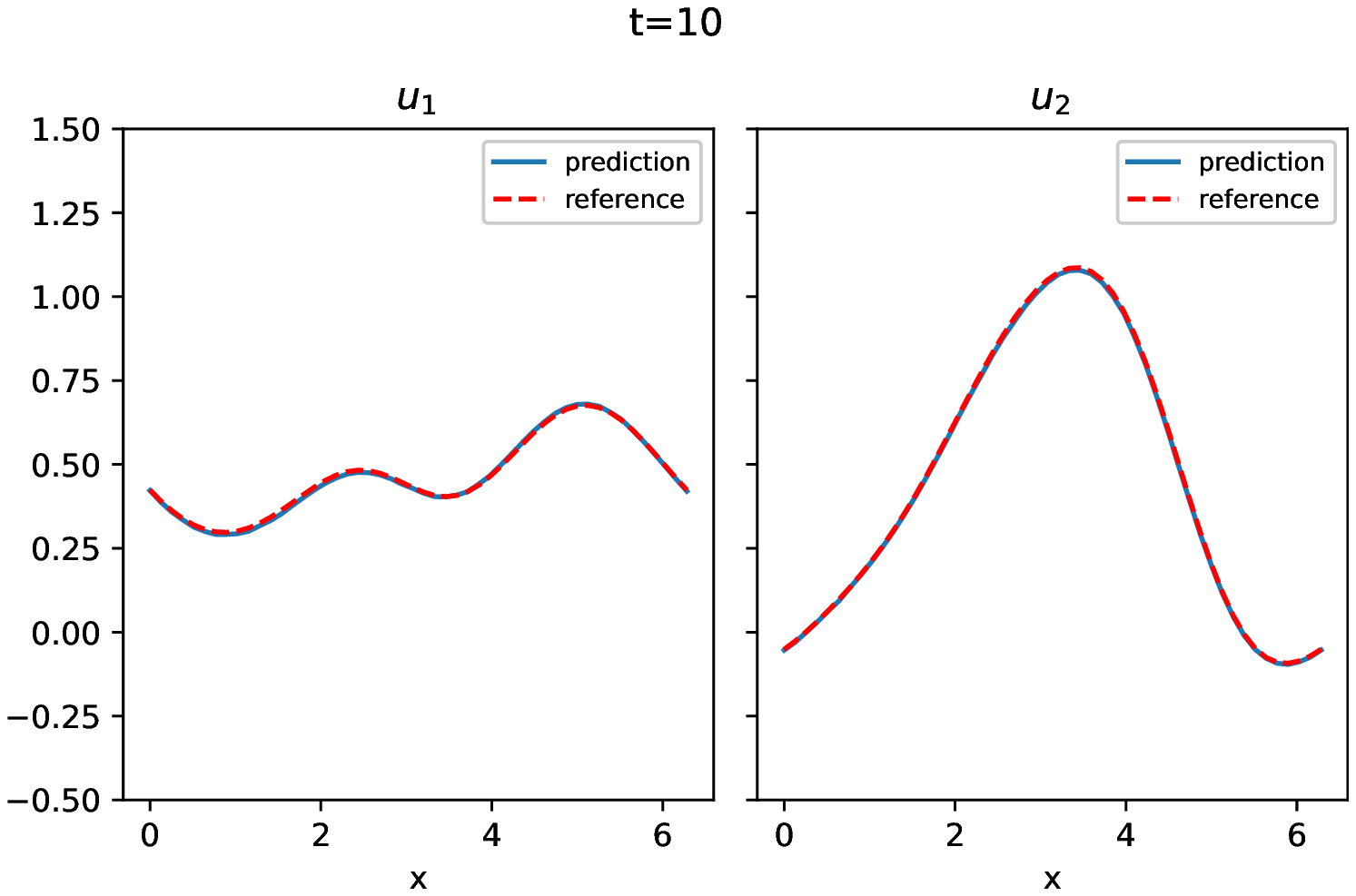}
		\caption{Systems of wave
                  equations. Comparison between DNN model prediction and reference solution.}
		\label{fig:ex6_system}
	\end{center}
\end{figure}

A set of prediction results generated by the trained DNN model are shown
in Fig. \ref{fig:ex6_system}. These are based on a new initial
condition
\begin{align*}
\mathbf{u}(x,0) &= \begin{bmatrix} \exp(\sin(x)) \\ \exp(\cos(x)) \end{bmatrix}.
\end{align*}
Compared to the true solution, we observe that the DNN model produces
accurate prediction for time up to $t=10$.

\subsection{Two Dimensional Example}

We now consider a two-dimensional  advection-diffusion equation
$$
\frac{\partial u}{\partial t} + \nabla \cdot (\boldsymbol{\alpha} u) = \nabla \cdot (\kappa \nabla u)
$$
on an unstructured grid over domain $(x,y)\in [-1,1]\times [-1,1]$ with zero Dirichlet boundary condition.
The transport velocity field is set as
$\boldsymbol{\alpha}(x,y) = (y, -x)^{T}$, and the viscosity is set as  $\kappa = 5 \times 10^{-3}$. 

\begin{figure}[htbp]
	\begin{center}
		\includegraphics[width=0.49\textwidth]{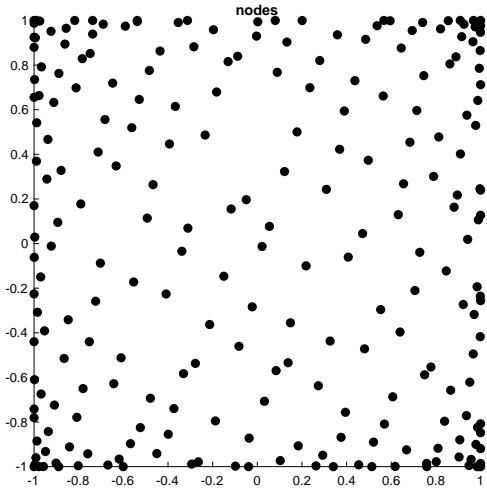}
		\caption{2D advection-diffusion. Unstructured grids.}
		\label{fig:ex6_nodes}
	\end{center}
\end{figure}

The unstructured grids are shown in Figure \ref{fig:ex6_nodes}. They consist of
200 points in the interior $(-1,1)\times (-1,1)$, 8 points along each edge (resulting 32 points over the edges), and the 4 corner points.
The interior points are generated using 2D Sobol sequence in $[0,\pi]\times [0,\pi]$, followed by cosine transformation.
The edge points are generated by uniform random distribution in $(0,\pi)$, followed by cosine transformation.

We generate $100,000$ trajectory data for training using initial condition
$$
u_0(x,y) = \sum_{k=1}^{N_c} \sum_{l=1}^{N_c} c_{k,l} \sin\left(\frac{k\pi}{2}(x+1)\right)\sin\left(\frac{l\pi}{2}(y+1)\right), 
$$
where $N_c=7$, and the coefficient $c_{k,l}^{(i)} \sim \frac{1}{k+l} U[-1,1]$. The training trajectory data are of length $n_L=10$ with time step $\Delta t=2\times 10^{-3}$, resulting training data of length $t=0.02$.

The DNN has disassembly block with dimension $236\times 1\times 3$ and assembly layer of dimension $1\times 1 \times 3$. The tanh activation function is used throughout, along with a constant learning rate of $1 \times 10^{-3}$. The DNN model is trained for 6,000 epochs with batch size 100.

\begin{figure}[htbp]
	\begin{center}
		\includegraphics[width=0.49\textwidth]{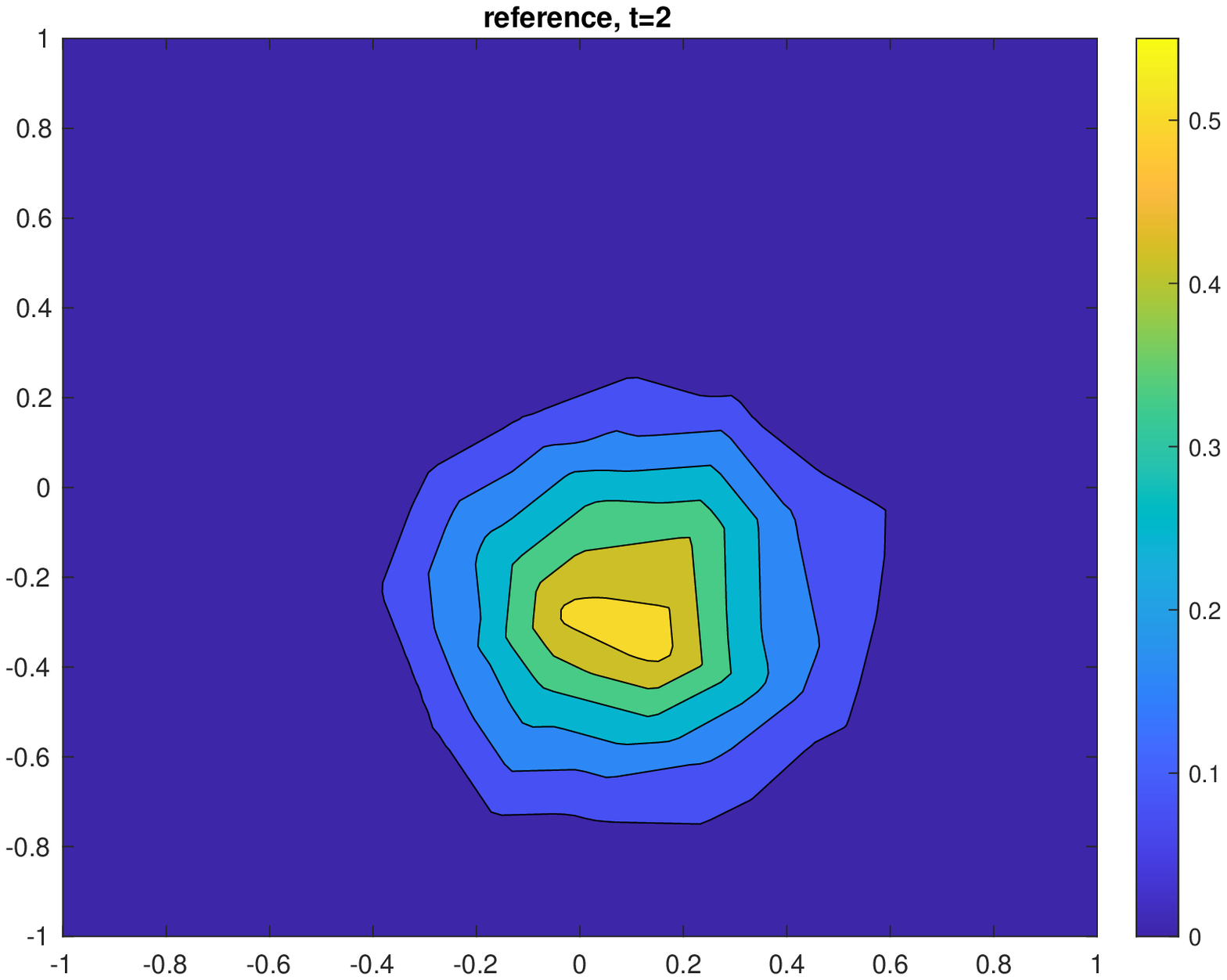}
		\includegraphics[width=0.49\textwidth]{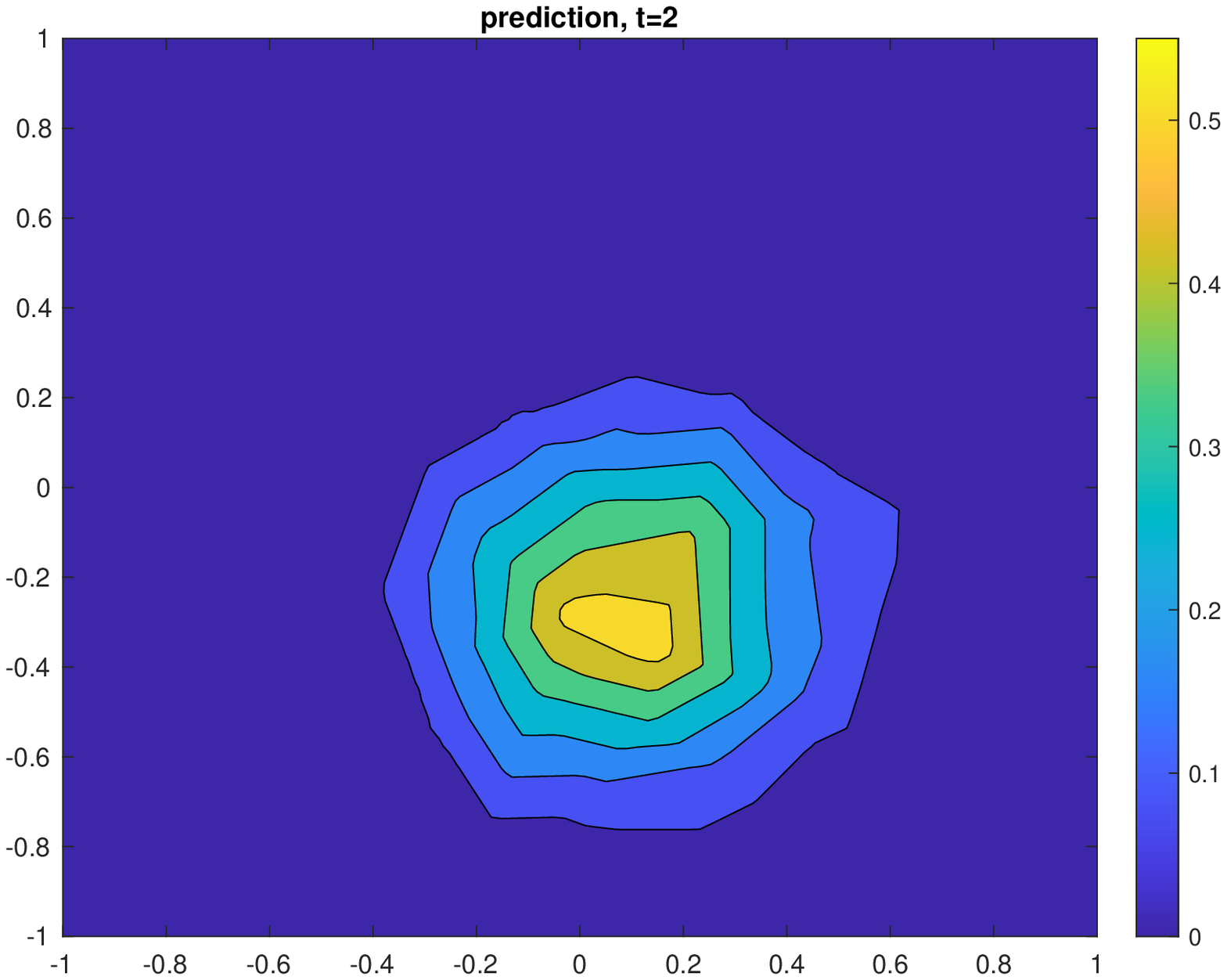}
		\includegraphics[width=0.49\textwidth]{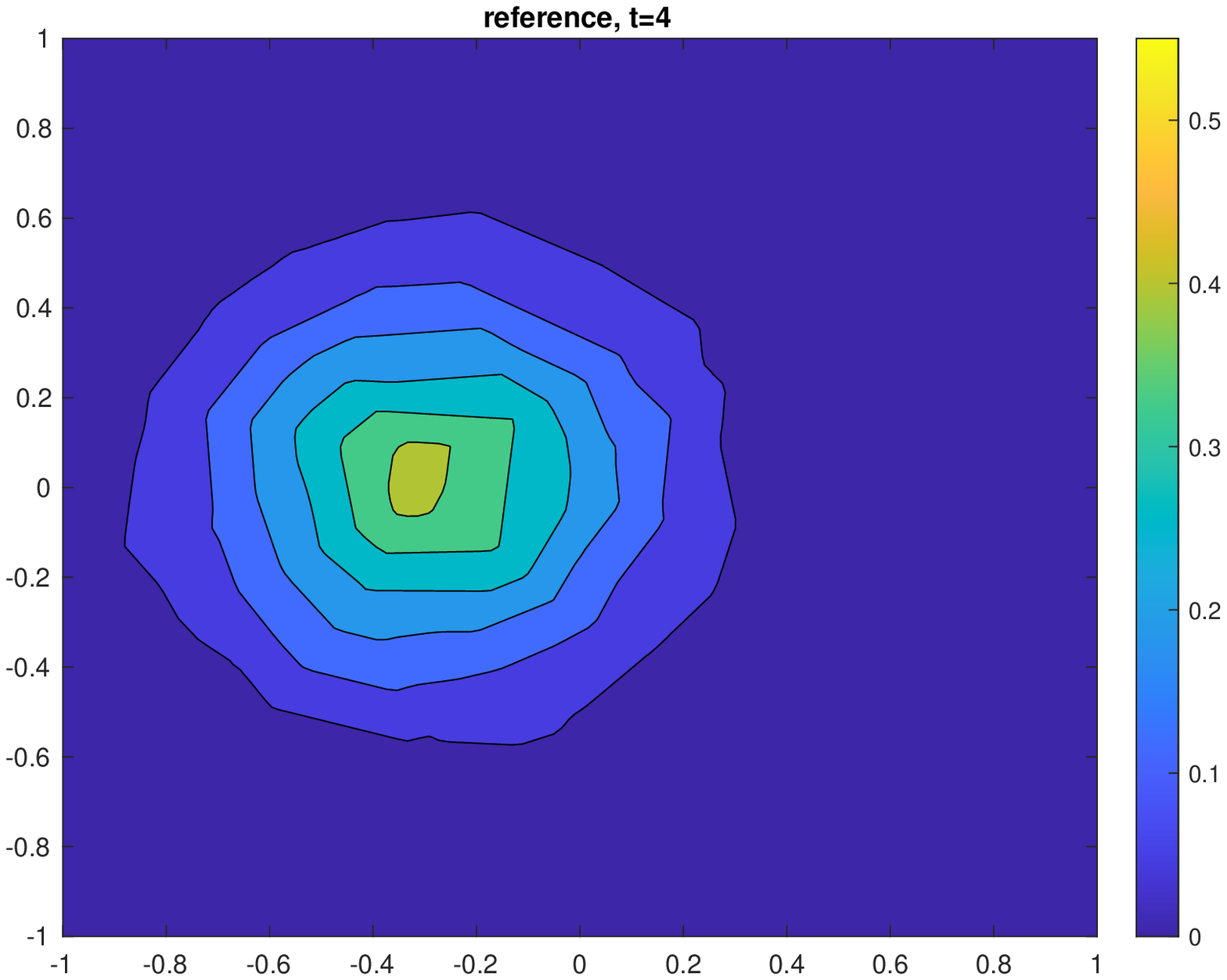}
		\includegraphics[width=0.49\textwidth]{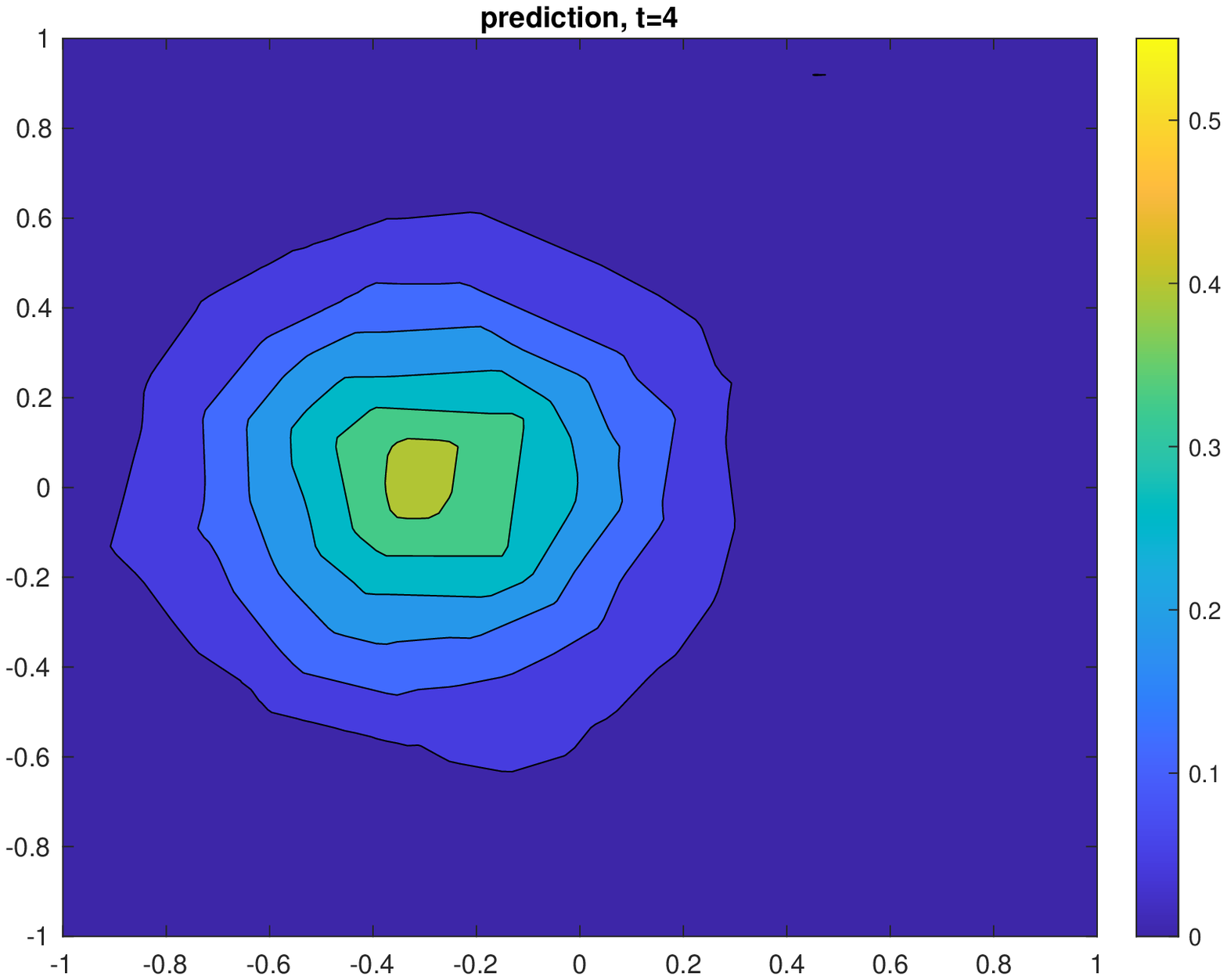}
		\caption{2D advection-diffusion over unstructured grids: Comparison of DNN model prediction (left column) and reference solution (right column).}
		\label{fig:ex6_contour}
	\end{center}
\end{figure}

For validation, we present the DNN prediction with a new initial condition 
$$
u_0(x,y) = \frac{C}{\sqrt{4\pi^2\sigma_x^2\sigma_y^2}} \exp\left[-\frac{1}{2}\frac{(x-\mu_x)^2}{\sigma_x^2}-\frac{1}{2}\frac{(y-\mu_y)^2}{\sigma_y^2}\right],
$$ 
where $C=0.2$, $\mu_x=\mu_y=0.2$, and $\sigma_x=\sigma_y=0.18$. In Figure \ref{fig:ex6_contour}, we plot the solutions at different time.. We observe good agreement between the DNN model prediction and the reference solution for up to $t=4$.

\section{Conclusion} \label{sec:conclusions}

We present a general framework for DNN learning of unknown PDEs
using nodal values of data. A new DNN structure is developed, and its
existence
is established mathematically. An extensive set of examples are
presented and demonstrate that the proposed DNN approach is able to model
a variety of model PDE problems from data, particularly unstructured
data. The current work introduces an alternative direction for data
driven modeling for PDE related problems.
More in-depth work on the proposed DNNs
will be pursued in a future work.

\bibliographystyle{siamplain}
\bibliography{neural,LearningEqs}

\begin{thebibliography}{10}

\bibitem{brunton2016discovering}
{\sc S.~L. Brunton, J.~L. Proctor, and J.~N. Kutz}, {\em Discovering governing
  equations from data by sparse identification of nonlinear dynamical systems},
  Proc. Natl. Acad. Sci. U.S.A., 113 (2016), pp.~3932--3937.

\bibitem{FuChangXiu_JMLMC20}
{\sc X.~Fu, L.-B. Chang, and D.~Xiu}, {\em Learning reduced systems via deep
  neural networks with memory}, J. Machine Learning Model. Comput., 1 (2020),
  pp.~97--118.

\bibitem{he2016deep}
{\sc K.~He, X.~Zhang, S.~Ren, and J.~Sun}, {\em Deep residual learning for
  image recognition}, in Proceedings of the IEEE conference on computer vision
  and pattern recognition, 2016, pp.~770--778.

\bibitem{kang2019ident}
{\sc S.~H. Kang, W.~Liao, and Y.~Liu}, {\em {IDENT}: Identifying differential
  equations with numerical time evolution}, arXiv preprint arXiv:1904.03538,
  (2019).

\bibitem{long2018pde}
{\sc Z.~Long, Y.~Lu, and B.~Dong}, {\em {PDE-Net} 2.0: Learning {PDEs} from
  data with a numeric-symbolic hybrid deep network}, arXiv preprint
  arXiv:1812.04426,  (2018).

\bibitem{long2017pde}
{\sc Z.~Long, Y.~Lu, X.~Ma, and B.~Dong}, {\em {PDE}-net: Learning {PDE}s from
  data}, in Proceedings of the 35th International Conference on Machine
  Learning, J.~Dy and A.~Krause, eds., vol.~80 of Proceedings of Machine
  Learning Research, Stockholmsm\"assan, Stockholm Sweden, 10--15 Jul 2018,
  PMLR, pp.~3208--3216.

\bibitem{QinChenJakemanXiu_IJUQ}
{\sc T.~Qin, Z.~Chen, J.~Jakeman, and D.~Xiu}, {\em Deep learning of
  parameterized equations with applications to uncertainty quantification},
  Int. J. Uncertainty Quantification,  (2020),
  p.~10.1615/Int.J.UncertaintyQuantification.2020034123.

\bibitem{QinChenJakemanXiu_SISC}
{\sc T.~Qin, Z.~Chen, J.~Jakeman, and D.~Xiu}, {\em Data-driven learning of
  non-autonomous systems}, SIAM J. Sci. Comput.,  (2021), p.~in press.

\bibitem{qin2018data}
{\sc T.~Qin, K.~Wu, and D.~Xiu}, {\em Data driven governing equations
  approximation using deep neural networks}, J. Comput. Phys., 395 (2019),
  pp.~620 -- 635.

\bibitem{raissi2018deep}
{\sc M.~Raissi}, {\em Deep hidden physics models: {Deep} learning of nonlinear
  partial differential equations}, Journal of Machine Learning Research, 19
  (2018), pp.~1--24.

\bibitem{raissi2017physics1}
{\sc M.~Raissi, P.~Perdikaris, and G.~E. Karniadakis}, {\em Physics informed
  deep learning (part i): Data-driven solutions of nonlinear partial
  differential equations}, arXiv preprint arXiv:1711.10561,  (2017).

\bibitem{raissi2017physics2}
{\sc M.~Raissi, P.~Perdikaris, and G.~E. Karniadakis}, {\em Physics informed
  deep learning (part ii): Data-driven discovery of nonlinear partial
  differential equations}, arXiv preprint arXiv:1711.10566,  (2017).

\bibitem{raissi2018multistep}
{\sc M.~Raissi, P.~Perdikaris, and G.~E. Karniadakis}, {\em Multistep neural
  networks for data-driven discovery of nonlinear dynamical systems}, arXiv
  preprint arXiv:1801.01236,  (2018).

\bibitem{rudy2017data}
{\sc S.~H. Rudy, S.~L. Brunton, J.~L. Proctor, and J.~N. Kutz}, {\em
  Data-driven discovery of partial differential equations}, Science Advances, 3
  (2017), p.~e1602614.

\bibitem{rudy2018deep}
{\sc S.~H. Rudy, J.~N. Kutz, and S.~L. Brunton}, {\em Deep learning of dynamics
  and signal-noise decomposition with time-stepping constraints}, J. Comput.
  Phys., 396 (2019), pp.~483--506.

\bibitem{scarselli1998universal}
{\sc F.~Scarselli and A.~C. Tsoi}, {\em Universal approximation using
  feedforward neural networks: A survey of some existing methods, and some new
  results}, Neural networks, 11 (1998), pp.~15--37.

\bibitem{schaeffer2017learning}
{\sc H.~Schaeffer}, {\em Learning partial differential equations via data
  discovery and sparse optimization}, Proceedings of the Royal Society of
  London A: Mathematical, Physical and Engineering Sciences, 473 (2017).

\bibitem{schaeffer2017sparse}
{\sc H.~Schaeffer and S.~G. McCalla}, {\em Sparse model selection via integral
  terms}, Phys. Rev. E, 96 (2017), p.~023302.

\bibitem{schaeffer2017extracting}
{\sc H.~Schaeffer, G.~Tran, and R.~Ward}, {\em Extracting sparse
  high-dimensional dynamics from limited data}, SIAM Journal on Applied
  Mathematics, 78 (2018), pp.~3279--3295.

\bibitem{sun2019neupde}
{\sc Y.~Sun, L.~Zhang, and H.~Schaeffer}, {\em {NeuPDE}: Neural network based
  ordinary and partial differential equations for modeling time-dependent
  data}, arXiv preprint arXiv:1908.03190,  (2019).

\bibitem{tran2017exact}
{\sc G.~Tran and R.~Ward}, {\em Exact recovery of chaotic systems from highly
  corrupted data}, Multiscale Model. Simul., 15 (2017), pp.~1108--1129.

\bibitem{WuQinXiu2019}
{\sc K.~Wu, T.~Qin, and D.~Xiu}, {\em Structure-preserving method for
  reconstructing unknown hamiltonian systems from trajectory data}, arXiv
  preprint arXiv:1905.10396,  (2019).

\bibitem{WuXiu_JCPEQ18}
{\sc K.~Wu and D.~Xiu}, {\em Numerical aspects for approximating governing
  equations using data}, J. Comput. Phys., 384 (2019), pp.~200--221.

\bibitem{WuXiu_modalPDE}
{\sc K.~Wu and D.~Xiu}, {\em Data-driven deep learning of partial differential
  equations in modal space}, J. Comput. Phys., 408 (2020), p.~109307.

\end{thebibliography}

\end{document}